\documentclass[lettersize,journal]{IEEEtran}
\usepackage{amsmath,amsfonts}
\usepackage{algorithm}
\usepackage{array}
\usepackage[caption=false,font=normalsize,labelfont=sf,textfont=sf]{subfig}
\usepackage{textcomp}
\usepackage{stfloats}
\usepackage{url}
\usepackage{verbatim}
\usepackage{graphicx}
\usepackage{cite}
\usepackage{multirow}
\usepackage{fancyhdr}
\usepackage{amsmath}
\usepackage{graphicx}
\usepackage{epstopdf}
\usepackage{calligra}
\usepackage{algorithmicx}
\usepackage{algpseudocode} 
\usepackage{soul}
\usepackage{color, xcolor} 

\usepackage{algorithm}
\usepackage{algpseudocode}

\errorcontextlines\maxdimen
\makeatletter
\newcommand*{\algrule}[1][\algorithmicindent]{\makebox[#1][l]{\hspace*{.5em}\thealgruleextra\vrule height \thealgruleheight depth \thealgruledepth}}%
\newcommand*{\thealgruleextra}{}
\newcommand*{\thealgruleheight}{.75\baselineskip}
\newcommand*{\thealgruledepth}{.25\baselineskip}

\newcount\ALG@printindent@tempcnta
\def\ALG@printindent{%
	\ifnum \theALG@nested>0
	\ifx\ALG@text\ALG@x@notext
	\else
	\unskip
	\addvspace{-1pt}
	\ALG@printindent@tempcnta=1
	\loop
	\algrule[\csname ALG@ind@\the\ALG@printindent@tempcnta\endcsname]%
	\advance \ALG@printindent@tempcnta 1
	\ifnum \ALG@printindent@tempcnta<\numexpr\theALG@nested+1\relax
	\repeat
	\fi
	\fi
}%
\usepackage{etoolbox}
\patchcmd{\ALG@doentity}{\noindent\hskip\ALG@tlm}{\ALG@printindent}{}{\errmessage{failed to patch}}
\makeatother

\newbox\statebox
\newcommand{\myState}[1]{%
	\setbox\statebox=\vbox{#1}%
	\edef\thealgruleheight{\dimexpr \the\ht\statebox+1pt\relax}%
	\edef\thealgruledepth{\dimexpr \the\dp\statebox+1pt\relax}%
	\ifdim\thealgruleheight<.75\baselineskip
	\def\thealgruleheight{\dimexpr .75\baselineskip+1pt\relax}%
	\fi
	\ifdim\thealgruledepth<.25\baselineskip
	\def\thealgruledepth{\dimexpr .25\baselineskip+1pt\relax}%
	\fi
	\State #1%
	\def\thealgruleheight{\dimexpr .75\baselineskip+1pt\relax}%
	\def\thealgruledepth{\dimexpr .25\baselineskip+1pt\relax}%
}

\usepackage{soul}
\begin{document}

\title{A Phone-based Distributed Ambient Temperature Measurement System with An Efficient Label-free Automated Training Strategy}

\author{Dayin~Chen,
        Xiaodan~Shi{\IEEEauthorrefmark{1}},
        Haoran~Zhang,~\IEEEmembership{Senior Member,~IEEE,}       Xuan~Song{\IEEEauthorrefmark{1}},~\IEEEmembership{Member,~IEEE,}
        Dongxiao~Zhang, 
        Yuntian~Chen,~\IEEEmembership{Member,~IEEE,}
        and Jinyue~Yan{\IEEEauthorrefmark{1}}

\thanks{* Xiaodan Shi, Xuan Song and Jinyue Yan are the corresponding authors.}
\thanks{Dayin Chen is with Department of Building Environment and Energy Engineering, The Hong Kong Polytechnic University, Hong Kong SAR, China; International Centre of Urban Energy Nexus, The Hong Kong Polytechnic University,  Hong Kong SAR, China; Research Institute for Smart Energy, The Hong Kong Polytechnic University, Hong Kong SAR, China; Eastern Institute for Advanced Study, Eastern Institute of Technology, Ningbo, China. E-mail: 23038748r@connect.polyu.hk}
\thanks{Xiaodan Shi is with School of Business, Society and Technology, Mälardalens University, 72123 Västerås, Sweden; Center for Spatial Information Science, the University of Tokyo, 5-1-5 Kashiwanoha, Kashiwa-shi, Chiba 277-8568, Japan. E-mail: xiaodan.shi@mdu.se}
\thanks{Haoran Zhang is with School of Urban Planning and Design, Peking University, Shenzhen, China. E-mail: h.zhang@pku.edu.cn}
\thanks{Xuan Song is with School of Artificial Intelligence, Jilin University, Changchun, China; Research Institute of Trustworthy Autonomous Systems, Southern University of Science and Technology (SUSTech), Shenzhen, China. E-mail: songx@sustech.edu.cn}
\thanks{Dongxiao Zhang and Yuntian Chen are with Ningbo Institute of Digital Twin, Eastern Institute of Technology, Ningbo, China. E-mail: dzhang@eitech.edu.cn; ychen@eitech.edu.cn}
\thanks{Jinyue Yan is with Department of Building Environment and Energy Engineering, The Hong Kong Polytechnic University, Hong Kong SAR, China; International Centre of Urban Energy Nexus, The Hong Kong Polytechnic University,  Hong Kong SAR, China; Research Institute for Smart Energy, The Hong Kong Polytechnic University, Hong Kong SAR, China. E-mail: j-jerry.yan@polyu.edu.hk}

}




\maketitle
\begin{abstract}
Enhancing the energy efficiency of buildings significantly relies on monitoring indoor ambient temperature. The potential limitations of conventional temperature measurement techniques, together with the omnipresence of smartphones, have redirected researchers’ attention towards the exploration of phone-based ambient temperature estimation methods. However, existing phone-based methods face challenges such as insufficient privacy protection, difficulty in adapting models to various phones, and hurdles in obtaining enough labeled training data. In this study, we propose a distributed phone-based ambient temperature estimation system which enables collaboration among multiple phones to accurately measure the ambient temperature in different areas of an indoor space. This system also provides an efficient, cost-effective approach with a few-shot meta-learning module and an automated label generation module. It shows that with just 5 new training data points, the temperature estimation model can adapt to a new phone and reach a good performance. Moreover, the system uses crowdsourcing to generate accurate labels for all newly collected training data, significantly reducing costs. Additionally, we highlight the potential of incorporating federated learning into our system to enhance privacy protection. We believe this study can advance the practical application of phone-based ambient temperature measurement, facilitating energy-saving efforts in buildings.

\end{abstract}


\section{Introduction}

\IEEEPARstart{M}{onitoring} indoor temperature is significant for energy saving in building systems \cite{hong2006impact}. The indoor thermal environment significantly influences occupants' comfort, well-being, and productivity, while also playing a crucial role in the energy consumption of residential, commercial, and industrial sectors \cite{YAN1}. By closely monitoring indoor temperature, we can identify potential inefficiencies, and implement targeted measures to optimize energy usage. Numerous studies \cite{energy1,energy2,energy3} have highlighted the importance of monitoring and controlling indoor temperature in achieving energy efficiency. 

Actually, in a large indoor space such as shopping malls, public offices and stadiums, the temperature is not always the same in different small areas. We think that accurately monitoring the temperature of different areas serves as the foundation for developing advanced technologies aimed at enhancing human comfort and energy conservation. For example, the traditional centralized cooling system can be transformed into a fine-grained distributed cooling system. By considering the number of occupants and the temperature estimation results, the cooling intensity in specific areas can be adjusted to optimize energy usage and conserve energy. Besides, personalized micro thermal comfort zones can be created leveraging fine-grained temperature monitoring. Instead of maintaining thermal comfort throughout the entire room, we can construct some small thermal comfort zones for each personal by applying mini-watt mobile cooling devices or implementing the distributed cooling on furniture such as desks or chairs. This approach ensures individualized comfort while also promoting energy efficiency.



The conventional and most commonly employed method for measuring indoor temperature typically is using thermometers. However, it suffers from several drawbacks including fixed location, limited measuring range, relatively high cost and configuration error \cite{chen2023}. With the consideration of these, in this study, we propose a distributed phone-based ambient temperature estimation system to help effectively monitor the indoor ambient temperature.

\subsection{Estimating Ambient Temperature with Smartphones}

Using smartphones to estimate ambient temperature has gained attraction in recent years, owing to their widespread availability \cite{0,1,2,3,4}. Though temperature sensors were previously integrated into smartphones, studies have indicated that these estimations are influenced by various factors which bring significant measurement bias \cite{0}. These years, researchers have shifted their focus towards estimating ambient temperature based on phone battery temperature and other phone state data. Given the aforementioned contexts, we present a phone-based ambient temperature estimation model with a machine learning structure. This model utilizes phone battery data and various phone state features as input and is capable of estimating the current ambient temperature, while also generating the associated uncertainty of this estimation. 


\subsection{Crowdsourcing and Truth Inference}

In many prior studies, the integration of crowdsourcing techniques has been employed to enhance the accuracy of temperature measurement \cite{1,2,3,4}. Crowdsourcing is a data annotation method that involves assigning the same task to multiple participants, allowing them to provide answers. Because of the differences in cognitive level and the existence of malicious sabotage behavior, the truth inference technology is introduced to aggregate all responses and derive a final accurate result. The typical truth inference algorithms includes Mean \cite{cs_all}, Participant-mine Voting (PM) \cite{PM}, Confidence-aware Truth Discovery (CATD) \cite{CATD}, ZenCrowd (ZC) \cite{ZC}, among others. These algorithms take into account different factors such as task difficulty, task domain, and participant reliability to infer the true value from multiple answers. 

In recent years, there has been a notable integration of uncertainty in reported answers within truth inference algorithms, as evidenced by the works \cite{uncertainty1,uncertainty2,uncertainty3,uncertainty4,uncertainty5}. However, despite the advancements made, these approaches still possess certain limitations. Firstly, these methods rely on self-reported uncertainty values of each answer as input. Nevertheless, in most crowdsourcing scenarios, organizers do not require participants to provide the uncertainty of their answers. This is mainly due to the abstract nature of uncertainty and the difficulty in expressing it. Furthermore, self-reported uncertainty measures are highly subjective and their reliability cannot be guaranteed. Additionally, the performance of these algorithms is particularly sensitive to the initial parameter values chosen. 

This study introduces a combined approach consisting of an automated uncertainty generation module and a truth inference module to aggregate multiple temperature estimation answers from diverse phones, while considering the uncertainty associated with each answer. This truth inference method, named Confidence-based Tree-structure (CBTS) model, effectively addresses the aforementioned limitations. It is initially introduced in our previous work \cite{chen2023}. However, in this study, we present a more comprehensive analysis and explore additional applications of this model. In the CBTS model, the uncertainty value used is automatically and passively generated by the temperature estimation model based on relevant data from the phone's state. Besides, It does not depend on initial parameters setting. We also find the CBTS model exhibits higher robustness towards fluctuations in the number of devices involved. Even with a limited number of participants, it consistently demonstrates high performance levels.

In addition to using the CBTS model to aggregate multiple estimation answers, considering the high cost associated with acquiring labeled data for each new phone to train an estimation model, this study also introduces a data annotation approach that leverages crowdsourcing technology and the CBTS model for automatic label construction. The proposed approach significantly reduces costs while facilitating the extension of the temperature estimation model to a broader range of new mobile phones.

\subsection{Meta-learning and Few-shot Learning}

A notable challenge of phone-based ambient temperature measurement is the variability of equipment specifications among different smartphones. A temperature estimation model that performs well on one type of phone cannot be directly applied to a new type of phone. To address this issue, a viable strategy is to train a general source temperature estimation model that can be fine-tuned with some new training data for adapting on a new phone. This strategy is referred to as meta-learning. On the other hand, the short duration that a new phone remains within a specific area renders it impractical to obtain a substantial amount of training data. Therefore, alternative training strategies need to be utilized to develop a model that requires only a small quantity of new data and can be deployed rapidly. In this study, the few-shot learning is integrated into our system. In summary, we propose an new approach to address these challenges through the utilization of a meta-learning-based few-shot learning strategy. 

Meta-learning, also known as "learning to learn," involves training a model on a variety of tasks to enable it to adapt efficiently to similar tasks \cite{meta_all}. Unlike traditional machine learning methods that train with data samples, meta-learning learns from multiple predefined tasks. It has demonstrated substantial effectiveness in diverse domains \cite{unsuperviser_learning, MAML, hyperparameter_optimization}. On the other hand, few-shot learning (FSL) is a specific machine learning paradigm tailored to situations where only a limited amount of labeled data is available for training \cite{few-shot1}. Many recent advancements in FSL have embraced a meta-learning approach, such as the RNN memory-based FSL \cite{RNN,rnn2,rnn3} and the metric-based FSL \cite{metric}. In our system, we adopt the Model-Agnostic Meta-Learning (MAML) framework \cite{MAML}, which is a widely used meta-learning-based few-shot learning framework, to train the temperature estimation model for each smartphone. The core concept of MAML is to obtain optimal initialization parameters, denoted as $\theta$. It can facilitate rapid task optimization from $\theta$ through one or more gradient descent steps using a small volume of available data \cite{few-shot1}.


\subsection{Federated Learning}

Privacy security is also a challenge. The collaborative training process inevitably requires sharing data, which results in privacy concerns and general reluctance among individuals to participate in ambient temperature measurement tasks. To address this challenge, we propose the incorporation of Federated Learning (FL) into our system as a means to safeguard the privacy of smartphone users.

FL, initially introduced by Google in 2016 \cite{FL_google}, has emerged as a promising approach for developing machine learning models that leverage data from multiple parties while ensuring data privacy \cite{fl_all}. This study specifically focuses on horizontal federated learning, which is a subtype of FL suitable for scenarios where diverse devices share a common set of features but possess distinct data samples \cite{fl_all}. The primary emphasis of horizontal federated learning lies in security, ensuring the protection of data exchange among clients to mitigate the risk of privacy leakage \cite{fl2}. To this end, various strategies, including homomorphic encryption \cite{homomorphic_encryption}, differential privacy \cite{differentially_private}, and secure aggregation \cite{secure_aggregation}, are employed to safeguard data information.

The primary aim of this study is to design a distributed system for estimating ambient temperature. Consequently, the focus on efficiency and accuracy of federated learning is relatively limited. In this study, we just demonstrate a straightforward scenario of combining the federated learning into our system to avoid privacy disclosure. The cryptographic technique we use is homomorphic encryption, which allows computation to be performed directly on encrypted data, yielding an encrypted result that, when decrypted, matches the result of the computation performed on the plaintext data \mbox{\cite{HE}}.

\subsection{Contributions}


Despite extensive research on phone-based ambient temperature estimation with crowdsourcing, several drawbacks still exist in the existing works \cite{1,2,3,4}. Specifically, the works conducted by \cite{2} and \cite{3} focus on city-scale estimation with daily resolution. They do not include any phone pattern information except the battery temperature. Moreover, their estimation is based on the physical model, which has a relatively poor performance. The work presented by \cite{4} addresses the issue of granularity in both time and space. However, their use of the physical model also results in a high estimation error. Furthermore, their experiments were limited to only three phones with sufficient data, without discussing the generalization to new phones or thoroughly researching the crowdsourcing aspect. Although a relatively comprehensive system was proposed in \cite{1}, it also exhibits certain shortcomings. To begin with, the ambient temperature estimation in their system relies on features such as CPU utilization and network information, which are no longer available on the latest phones. Besides, their crowdsourcing method considers the uncertainty of each phone instead of each estimation result. However, since a single phone may provide both accurate and inaccurate estimations, assigning uncertainty to individual estimations would be a more reasonable approach. Additionally, the practical value of the system could be further enhanced by addressing aspects such as data collection for new phones and privacy protection, which were not adequately explored.

In this study, we present a distributed and secure phone-based system specifically designed for indoor ambient temperature measurement. Our system comprises four essential modules: the ambient temperature measurement module, which enables each phone to estimate air temperature using a machine learning model; the crowdsourcing module, which collects multiple estimation answers for the same area from diverse phones to generate a more accurate result; the label generation model, which automatically generates labels for newly collected data through crowdsourcing; and the few-shot learning model, which facilitates rapid acquisition of a functional estimation model by each newly joined phone. Additionally, we also highlight the potential of incorporating federated learning to ensure robust user privacy protection. As this study builds upon our previous work \cite{chen2023}, we have condensed the descriptions of the first and second modules in this paper to reduce the overall length of exposition.

Through the implementation of our system, we effectively address several challenges commonly associated with existing phone-based ambient temperature measurement technologies. In summary, the key contributions of our work can be summarized as follows:
\begin{enumerate}


\item{We present a distributed cooperative system for estimating ambient temperature using mobile phones, which is highly practical and efficient.
}

\item{We introduce a crowdsourcing-based approach for automatic data annotation in order to assist newly joined phones in building their data sets, which yields significant cost reductions. }

\item{
We propose a few-shot learning strategy which facilitates rapid training of temperature estimation models and requires only a few pieces of data. It expedites participation of new phones in temperature estimation tasks.
 }

\item{We demonstrate the viability of incorporating federated learning into the model training process, ensuring the protection of mobile phone users' privacy.}

\end{enumerate}









\section{Methodology}

\begin{figure*}
	\centering
		\includegraphics[scale=0.45]{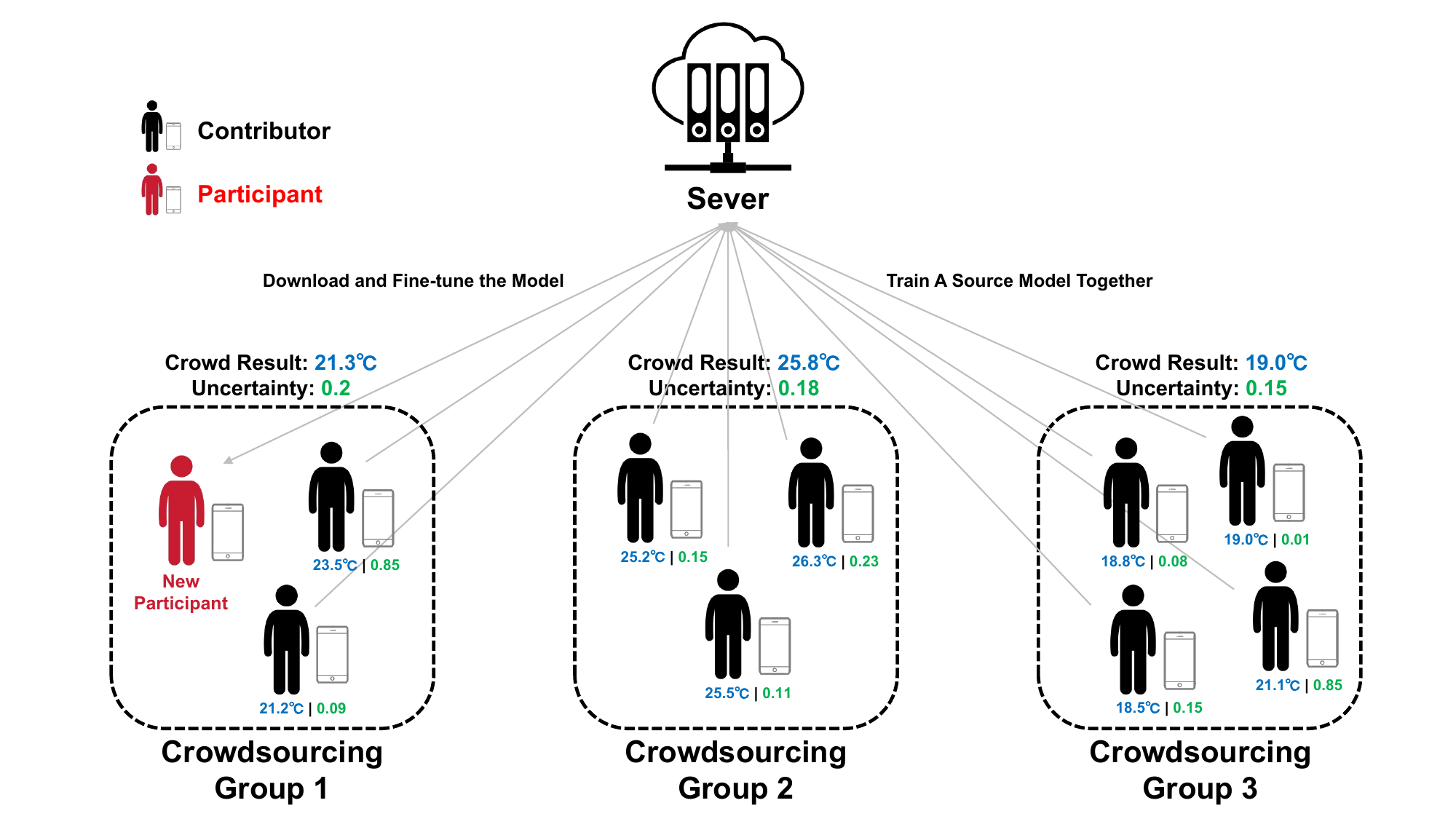}
	\caption{The schematic diagram of our distributed phone-based cooperative ambient temperature estimation system. 
}
	\label{framework1}
\end{figure*}

\subsection{System Overview}
\label{sys_overview}


We introduce our distributed phone-based cooperative ambient temperature estimation system through Fig. \ref{framework1}. In our system, we classify smartphone users into two categories:
\begin{itemize}
    \item {\textbf{Contributors}: 
    These users possess a precise temperature estimation model and can participate in training a source model in the server.
    }
    
    \item {\textbf{Participants}: These users are those who just joined the system. They lack both trained estimation models and labeled data. They do not contribute to the training of the source model. However, they can download the source model from the server and collect data to fine-tune a temperature estimation model for their own device.}
\end{itemize}

In our system, we assume the presence of contributors in every crowdsourcing group. All the contributors are from two sources: Some of the contributors are those who have manually collected a substantial amount of labeled data, enabling them to train an adequate model. Others are those phone users who are originally defined as the participant but have been in the system for a considerable duration and have refined their models to achieve high accuracy. This definition ensures a robust pool of contributors, as every participant has the potential to transition into a contributor after a certain period of time.

From Fig. \ref{framework1}, we can see that each contributor can provide an estimation of the ambient temperature. The blue numerical value represents the estimated temperature value, while the green value represents the uncertainty associated with this estimation. By utilizing the CBTS truth inference model to aggregate all the estimation answers within a crowdsourcing group, a more accurate answer can be obtained. The methodology of this part will be described in sections \ref{temp_model_section} and \ref{CBTS}. Additionally, all the contributors participate in the training of a source model using the MAML framework. The trained source model is stored on a central server. We will explain this part in section \ref{meta}. Every time a new participant joins a crowdsourcing group, it downloads the source model from the server and commence data collection together with other group members. The system incorporates an automated label generation module which can help assign labels to all the new collected data. The participant then uses these labeled data to establish a training data set and subsequently perform fine-tuning on the source model. Further elaboration on these procedures will be presented in section \ref{annotate}. Finally, in section \ref{fl}, the application of federated learning in this system will be demonstrated as a means to safeguard user privacy.

\subsection{Ambient Temperature Estimation Model}
\label{temp_model_section}

Our temperature estimation model incorporates a range of phone state features, including screen state, battery temperature, and others. All the features utilized in our model are listed in Table \ref{FEA}. The process of feature construction has been previously described in a separate article and is reiterated in the appendix of this paper. In comparison to related studies, these selected features possess a lower security level and can be acquired with relative ease.

\begin{table}[]
\center
\setlength{\tabcolsep}{5mm}
\caption{The collected features from smartphones.}
\label{FEA}
\begin{tabular}{ll}
\hline
Index & Collect Features\\ \hline
1 & Screen state\\
2& Battery Voltage \\
3 & Battery Temperature\\
4&Continuous screen activation time\\
5&Continuous screen off time\\
6&Before active, the continuous screen off time\\
7&Before off, the continuous screen active time\\
8&Final Battery temperature when screen was activated \\
9& Final Battery temperature when screen was turn off \\ \hline
\end{tabular}
\center
\end{table}

\begin{figure*}
	\centering
		\includegraphics[scale=0.45]{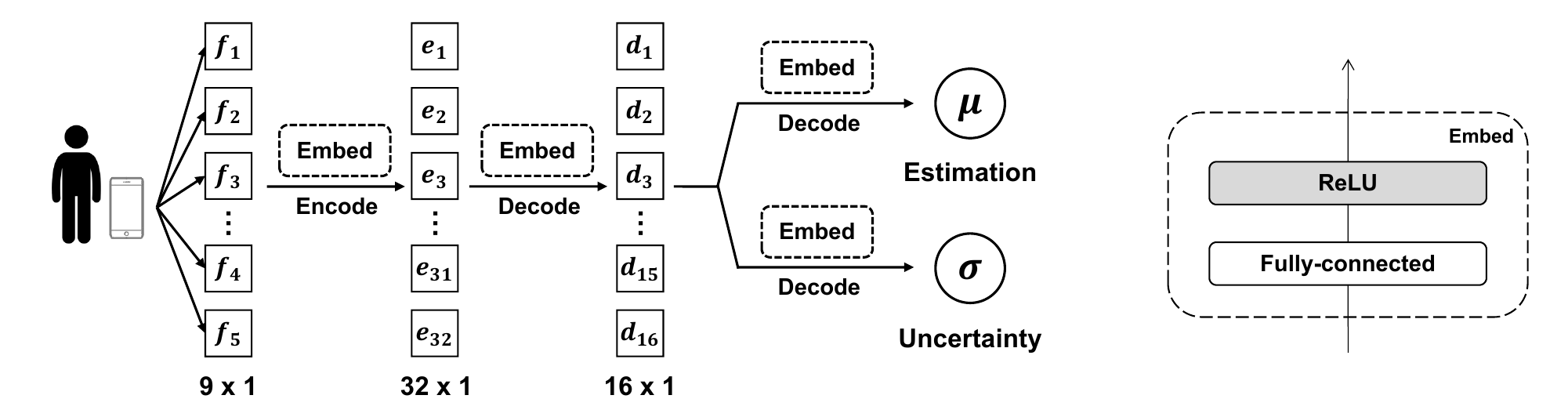}
	\caption{The structure of the ambient temperature estimation model. This model takes 9 phone state features as input and outputs the temperature estimation result with the corresponding uncertainty value.
}
	\label{temp_model}
\end{figure*}

The key innovation of this temperature estimation model lies in its ability to estimate a distribution rather than a specific numerical value for the ambient temperature. Fig. \ref{temp_model} provides a detailed structure of our model. To begin with, we encode the 9 phone state features into a 32-dimensional vector, which is then decoded back to 16 dimensions. This encoding and decoding process, referred to as \emph{Embed} for simplicity, comprises of a fully connected layer followed by a ReLU activation layer. At the final stage, the latent vector is decoded by two distinct \emph{Embed} operation modules in parallel. Both of these modules decode the latent vector into a numerical value. The first value represents the expected value of the predicted distribution and serves as a specific temperature estimation result. The second value represents the variance of the predicted distribution, thereby quantifying the uncertainty associated with the estimation result. We define the following loss function:
\begin{equation}
L(\mu, \sigma, T) = -log \frac{1}{\sqrt{2\pi}\sigma}e^{-\frac{(T-\mu)^2}{2\sigma^2}}
  \label{XX}      
\end{equation}
Here, T is the observed ambient temperature (ground truth), $\mu$ is the expectation value and $\sigma$ is variance. The supervision process entails maximizing the probability of the target value within the predicted Gaussian distribution, utilizing the output values of $\mu$ and $\sigma$. The appendices contain several concrete examples that illustrate the rationale behind the design of the loss function.




\subsection{CBTS Truth Inference Model}
\label{CBTS}

\begin{figure}
	\centering
		\includegraphics[scale=0.2]{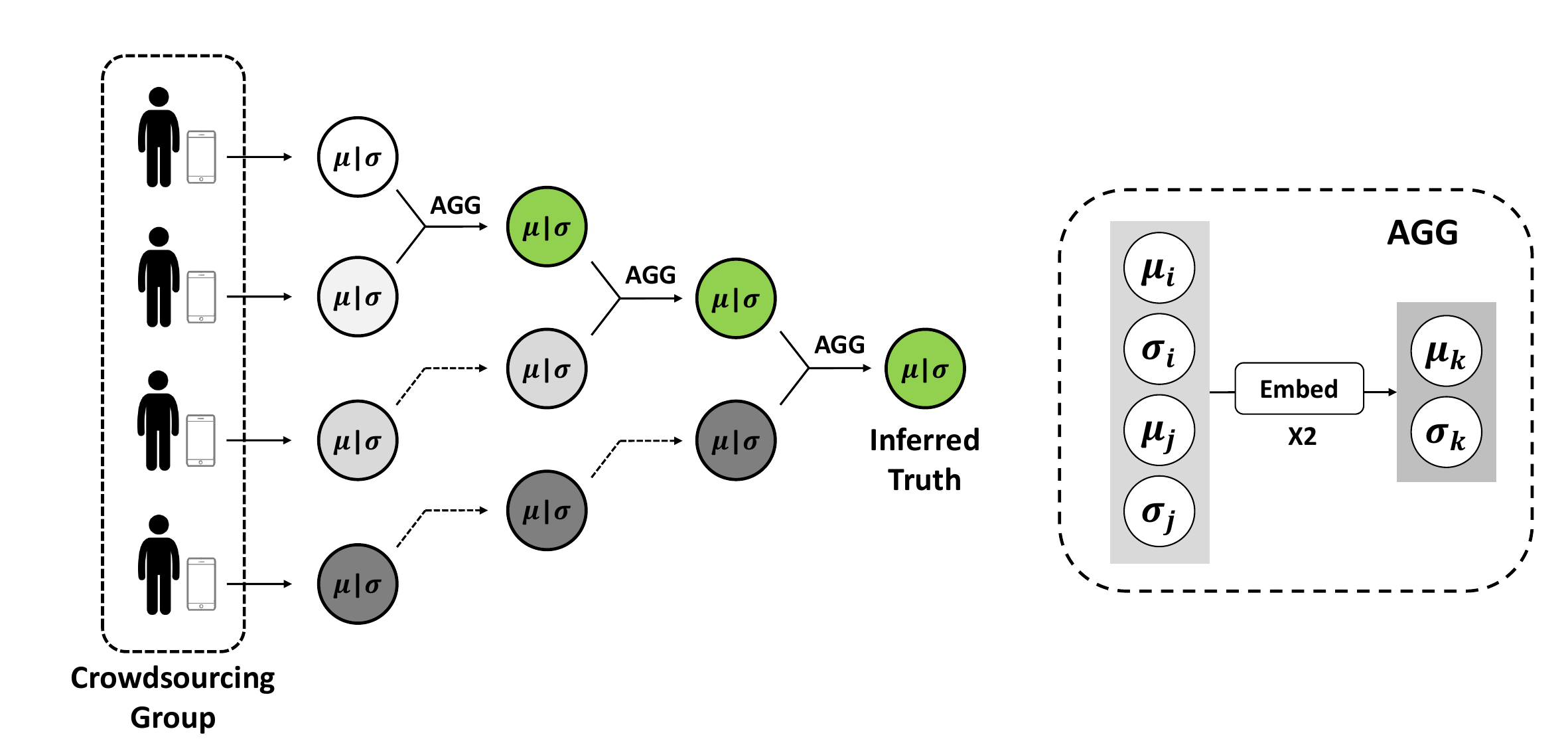}
	\caption{Structure of the CBTS truth inference model.}

	\label{cs_model}
\end{figure}

Truth inference models have been developed to leverage estimation results from multiple sources and generate a singular, accurate answer during crowdsourcing. In our system, we employ the CBTS model to do the truth inference, which is firstly introduced in our previous work \cite{chen2023}. The structure of the CBTS model is illustrated in Fig. \ref{cs_model}. For ease of illustration, we define an answer as a pair consisting of an estimated result ($\mu$) and its associated uncertainty value ($\sigma$), both derived from a single phone. The core operation in this model is termed \emph{Aggregate} (AGG), which combines two distinct answers to produce a new, singular answer. As depicted in Fig. \mbox{\ref{cs_model}}, AGG takes as input answer $i$ (comprising $\mu_i$ and $\sigma_i$) and answer $j$ (comprising $\mu_j$ and $\sigma_j$). Following the application of two Embed operations, AGG outputs a new answer $K$ (including $\mu_k$ and $\sigma_k$). Within a group of multiple given answers, the CBTS model continuously aggregates the current answer with the newly generated one until all answers in the group are combined, culminating in the final inferred answer.

In our system, we employ the CBTS model to fulfill two main objectives. Firstly, it aggregates multiple given answers within a crowdsourcing group to derive a more reliable estimation for one small area. Secondly, it generates inferred labels for new participants to aid them in constructing a training data set. Subsequently, participants can utilize the training data set to fine-tune the source model and obtain their personalized estimation models.

\subsection{Automatically Label Generation by Crowdsourcing}
\label{annotate}

\begin{figure}
	\centering
		\includegraphics[scale=0.26]{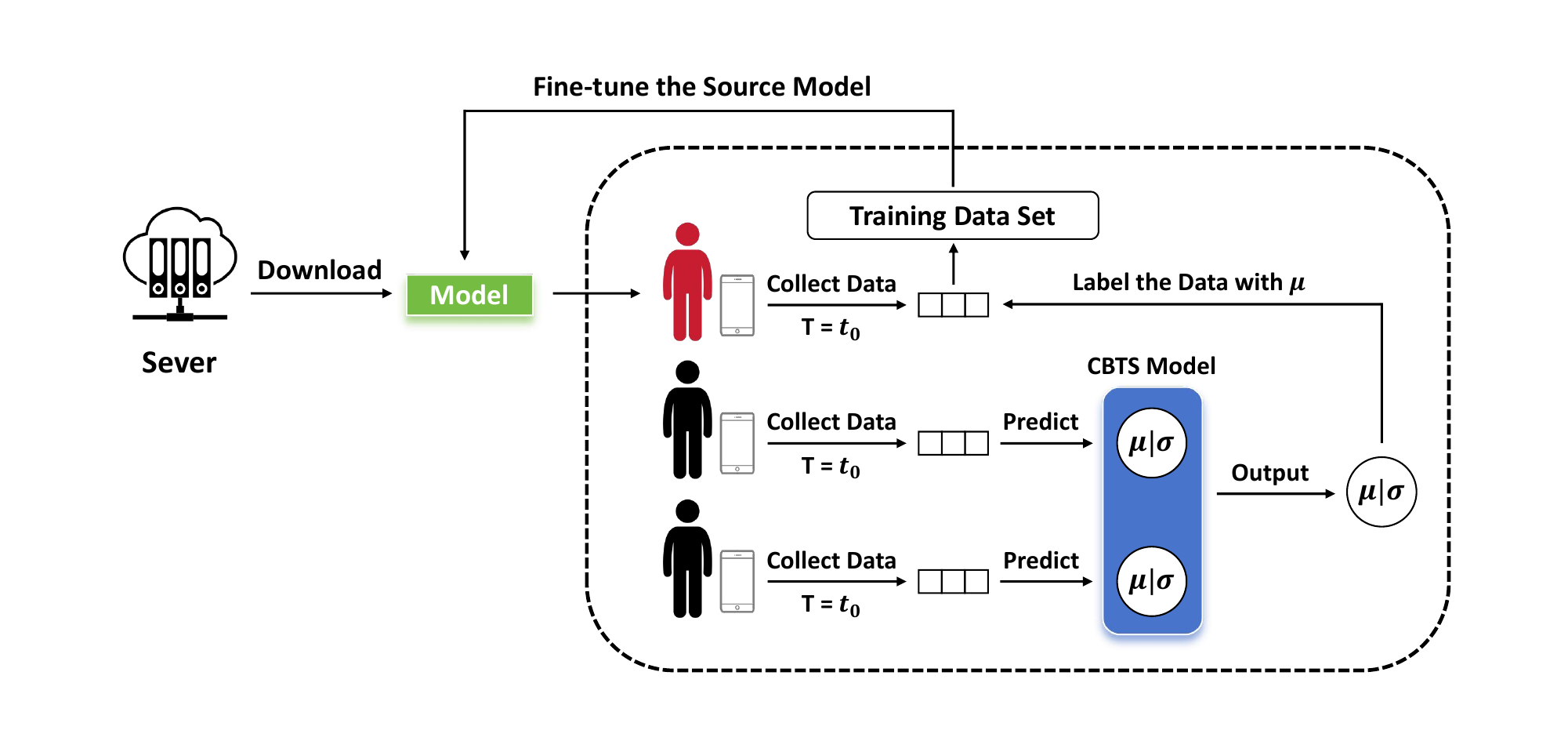}
	\caption{The flow diagram of how to build a training data set and fine-tune the source model for the new participant.
}
	\label{framework3}
\end{figure}

To address the challenges of limited training data and costly data annotations, we propose an automatic data annotation method that utilizes crowdsourcing technology to generate labeled data for each new participant. As depicted in Figure \ref{cs_model}, when a new participant joins a crowdsourcing group, it engages in data collection alongside other phones (contributors) within the group, operating at the same data acquisition frequency. Given that these contributors possess a trained temperature estimation model and can provide estimation answers based on the collected data, we employ the CBTS model to aggregate the estimation answers from all contributors and derive a final precise answer. This final answer is then considered as the inferred label for the corresponding data collected by the new participant. Through this approach, the need for manual data labeling is obviated, resulting in significant cost reduction. Furthermore, subsequent experimental results demonstrate that substituting the true label with the inferred label does not compromise the effectiveness of model training.


\subsection{Meta-learning-based Few-shot Learning}
\label{meta}

Given the diversity of hardware configurations and the transient nature of smartphones in fixed locations, we have incorporated the MAML framework into our system. This integration helps to effectively train a model utilizing a limited number of data samples and within constrained timelines. The forthcoming sections will provide a comprehensive explanation of the entire framework.

\subsubsection{Meta-Task}


\begin{figure}
	\centering
		\includegraphics[scale=0.2]{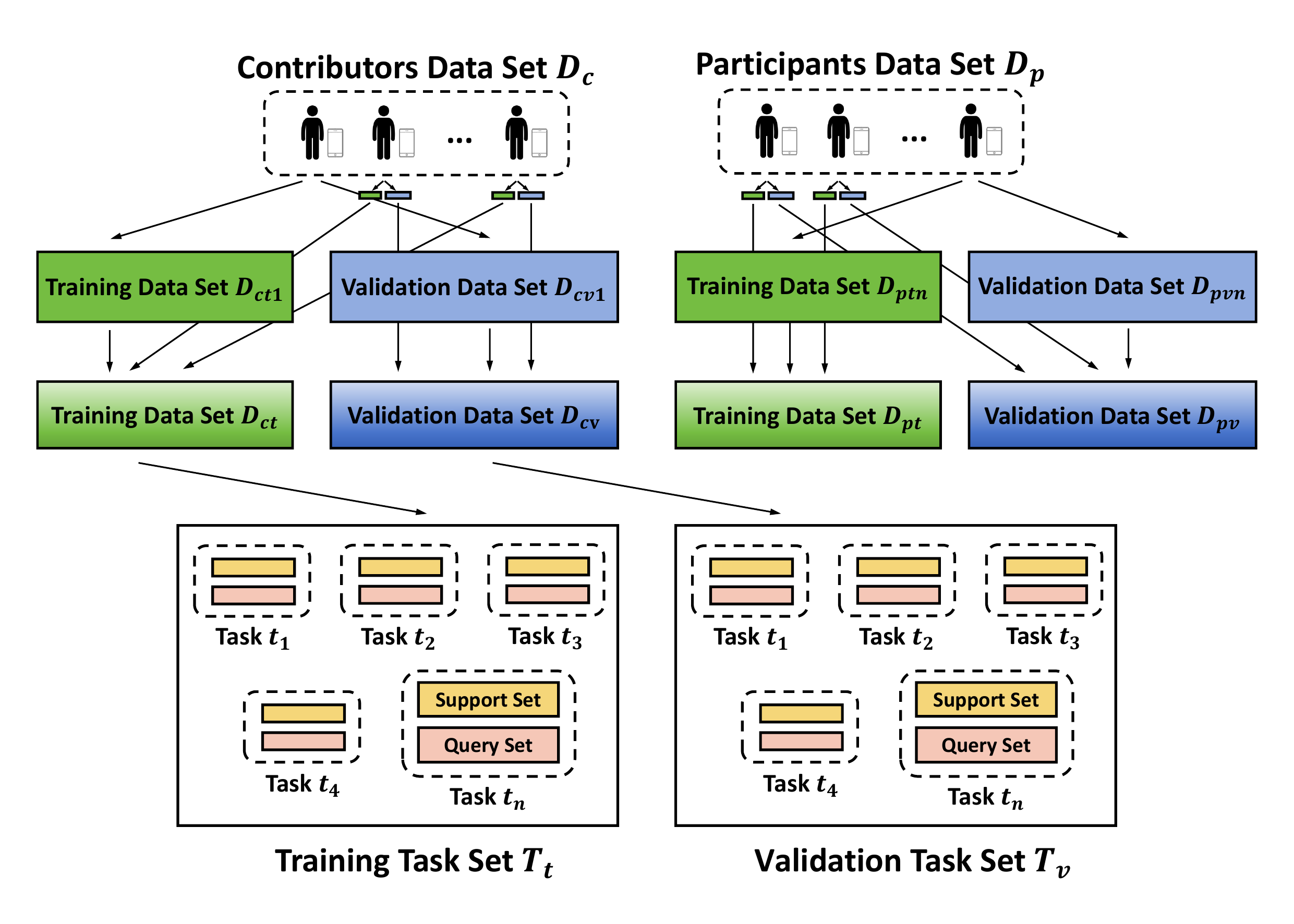}
	\caption{The diagram which illustrates the partition of data sets and task sets.}

	\label{dataset_fig}
\end{figure}

We utilize Fig. \ref{dataset_fig} to depict the dataset partitioning. As outlined in Section \ref{sys_overview}, we manually separate all the phones in our data set into two distinct groups: ``Contributors'' and ``Participants''. The data set corresponding to the ``Contributors'' and ``Participants'' groups are denoted by $\mathcal{D}_c$ and $\mathcal{D}_p$, respectively. Besides, the data of each contributor (denoted by $C_n$) in $\mathcal{D}_c$ is divided into two subsets: the training data set, $\mathcal{D}_{ctn}$, and the validation data set, $\mathcal{D}_{cvn}$ , with a ratio of 7:3. The combination of all contributor training data sets is denoted by $\mathcal{D}_{ct}$ and the combined validation data set of all contributors is denoted by $\mathcal{D}_{cv}$. The same data partitioning strategy is also employed for the participants data set, $\mathcal{D}_p$. In this context, all the aforementioned notations simply involve substituting the letter `c' with `p'.


Different from traditional training methods, the MAML framework necessitates training the model using predefined tasks rather than individual data samples. Within our system, we conceptualize a task as the process of ``Training a model for temperature estimation utilizing a support set, and subsequently evaluating its performance with a query set''. This conceptualization is in alignment with the foundational principles of MAML \mbox{\cite{MAML}}. Here, the support set plays a pivotal role in task-level model training, whereas the query set is instrumental in evaluation and updating the original model parameters. We denote the sizes of the support and query sets within a single task as $k_{spt}$ and $k_{qry}$, respectively. Importantly, all samples within a single task originate from the same phone. The details of the training methodology will be elaborated upon in the subsequent section \mbox{\ref{meta_train}}.

The remaining portion of Fig. \ref{dataset_fig} illustrates the construction of task sets. The figure demonstrates that we generate the training task set $\mathcal{T}_{t}$ using data from $\mathcal{D}_{ct}$, while the validation task set $\mathcal{T}_{v}$ is derived from data within $\mathcal{D}_{cv}$. Here, the training task set $\mathcal{T}_{t}$ is used to conduct the meta-training and the validation task set $\mathcal{T}_{v}$ is used to conduct the meta-evaluation. For each individual task, a random selection of $(k_{spt} + k_{qry})$ data samples is drawn from a single phone within the corresponding data set. Subsequently, $k_{spt}$ data samples are designated to compose the support set, while the remaining data samples are assembled to form the query set. Since we focus on few-shot learning, the value of $k_{spt}$ is very small.




\subsubsection{Meta-Training}
\label{meta_train}

The training methodology is concisely described in Algorithm \ref{algo1}. Initially, we set the task-level learning rate $\alpha$, meta-learning rate $\beta$, task count $n$, task-level update steps $s_1$, and the training task set $\mathcal{T}_{t}$. Each training epoch consists of the following sequence of operations: To begin with, a training batch is constructed by randomly selecting $n$ tasks from $\mathcal{T}_{t}$. Then, we make a copy of the original model parameters $\theta$ as $\theta^{'}_{i}$ and pick one task $t_i$ from the training batch. Subsequently, we take the loss value on support set of the task $t_i$ to update the copied parameters $\theta^{'}_{i}$ with the learning rate $\alpha$ for $s_1$ iterations. After that, we use the updated parameters $\theta^{'}_{i}$ to calculate the query loss value $\mathcal{L}_{qry}(f_{\theta^{'}_i})$ on the query set. This process is iteratively conducted for each task in the batch, accumulating the query set loss values. Upon completion of the batch processing, the aggregate of these query loss values is used to perform an overarching update on the original model parameters $\theta$ with the meta-learning rate $\beta$.


During the whole process, we continually take $n$ tasks from the training task set $\mathcal{T}_t$ to construct the training batch and repeat the above steps until $\mathcal{T}_t$ is empty.

\begin{algorithm}[ht!]  
	\renewcommand{\algorithmicrequire}{\textbf{Input:}}
	\renewcommand{\algorithmicensure}{\textbf{Output:}}
	\caption{Meta-Training}  
	\label{algo1}
	\begin{algorithmic} 
		\Require $\alpha$, $\beta$, $n$, $s_1$, $\mathcal{T}_t$ 
        \Repeat
            \State shuffle $\mathcal{T}_t$
            \Repeat
                \State build the training batch with $n$ task samples
                \For{task $t_{i}$, $i = 1, 2, ..., n$}
                    \State copy the parameters $\theta$ as $\theta^{'}_i$
                    \For{$step$ = 1, 2, ..., $s_1$} 
                    \myState {calculate $\mathcal{L}_{spt}(f_{\theta^{'}_i})$ with the support set}
                    \myState {update $\theta^{'}_i = \theta^{'}_i - \alpha\bigtriangledown_{\theta^{'}_i}\mathcal{L}_{spt}(f_{\theta^{'}_i})$}

                    \EndFor
                 \myState {calculate $\mathcal{L}_{qry}(f_{\theta^{'}_i})$ with the query set}
                \EndFor
                \myState {update $\theta = \theta - \beta\bigtriangledown_{\theta}\sum\limits_{i=1}\limits^{n_{t}}\mathcal{L}_{qry}(f_{\theta^{'}_i})$}
            \Until {end of $\mathcal{T}_t$}
        \Until {reach maximal epoch}
		
	\end{algorithmic}
\end{algorithm}

\subsubsection{Meta-Validation}

\begin{algorithm}[ht!]  
	\renewcommand{\algorithmicrequire}{\textbf{Input:}}
	\renewcommand{\algorithmicensure}{\textbf{Output:}}
	\caption{Meta-Validation}  
	\label{algo2}
	\begin{algorithmic} 
		\Require $\alpha$, $s_2$, $\mathcal{T}_v$ 
        \For{every task $t_{i}$ in $\mathcal{T}_v$}
            \State copy the parameters $\theta$ as $\theta^{'}_i$
            \For{$step$ = 1, 2, ..., $s_2$} 
                \myState {use data in the support set to calculate $\mathcal{L}_{spt}(f_{\theta^{'}_i})$}
                \myState {update $\theta^{'}_i = \theta^{'}_i - \alpha\bigtriangledown_{\theta^{'}_i}\mathcal{L}_{spt}(f_{\theta^{'}_i})$}
            \EndFor
            \myState {use $\theta^{'}_i$ the query set to evaluate model performance}
            
        \EndFor

	\end{algorithmic}
\end{algorithm}

Algorithm \ref{algo2} describes the meta-validation methodology, utilizing the task-level learning rate $\alpha$, number of fine-tuning steps $s_2$, and the validation task set $\mathcal{T}_v$ as inputs. In contrast to the training process, the validation process is considerably simpler. Initially, we define $s_2$ as the number of fine-tuning steps. For each task within the validation task set $\mathcal{T}_v$, we employ its support set to calculate the loss $\mathcal{L}_{spt}(f_{\theta^{'}_i})$ and subsequently update the model parameters $\theta^{'}_i$ accordingly. Following $s_2$ iterations of refinement, the task's query set is used to assess model performance.

\subsection{Federated Learning}
\label{fl}

During the meta-training process, we observe that the gradient of each task is computed independently. Each task copies the original model parameters and performs calculations using its own data set and does not interfere with each other. Additionally, the model parameters are updated by aggregating the gradients from all tasks. These characteristics render the approach highly suitable for federated learning applications.

In this study, we present a straightforward illustration of the application of federated learning in our system. The homomorphic encryption is employed to safeguard the privacy of phone users. To begin with, the server initializes a source model. Subsequently, each contributor downloads the source model and conducts local training using their respective data, thereby avoiding the need to upload their data to the server. This decentralized approach ensures the non-disclosure of sensitive data. Following the training phase, all phones utilize the public key to encrypt their respective gradients, which are then securely transmitted to the server through reliable communication channels. Since the homomorphic encryption technique is used, the server can aggregate all gradients in encrypted state without decryption. Upon the aggregation of all individual gradients, the server employs the private key to decrypt the aggregated result, thereby retrieving the meta-learning gradient. Finally, the server employs this gradient to update the source model parameters. Algorithm \mbox{\ref{algo3}} depicts the process of combining meta-learning and federated learning. We can find that the introduction of federated learning minimally alters the original MAML training process.

In this scenario, all personal data is stored locally, eliminating the need for phone users to share their data with other users or the server, thereby preserving their privacy. This approach alleviates privacy concerns and enhances user willingness to participate. Considering that gradients may inadvertently disclose user information, we propose a rule wherein the server performs decryption only after aggregating all encrypted gradients. Alternatively, the task of encrypted gradient aggregation can be delegated to a trusted third party. To maintain the article's structure, we defer the discussion of other attack scenarios to our future research. The specific homomorphic encryption method employed in this study is CKKS \mbox{\cite{CKKS}}. Due to the intricate mathematical nature of CKKS principles, we refrain from providing an in-depth explanation in this study.



\begin{algorithm}[ht!]  
    \renewcommand{\algorithmicensure}{\textbf{Server:}}
	\renewcommand{\algorithmicrequire}{\textbf{ClientCalculate($k$, $\theta$):}}

	\caption{Federated Learning}  
	\label{algo3}
	\begin{algorithmic} 
		\Ensure
        \State initialize $\theta$
        \For{each round $t$ = 1, 2, ...}
            \myState {send the model parameters $\theta$ to all clients in $\mathcal{K}$}
            \For{each client $k$ in $\mathcal{K}$}
                \myState{$E(g_k)$ = $ClientCalculate(k, \theta)$}
            \EndFor
            \myState {sum up encrypted gradients: $E(\sum\limits_{k=1}\limits^{\mathcal{K}}g_k)$ = $\sum\limits_{k=1}\limits^{\mathcal{K}}E(g_k)$}
            \myState {decrypt and get the result: $\sum\limits_{k=1}\limits^{\mathcal{K}}g_k = D(E(\sum\limits_{k=1}\limits^{\mathcal{K}}g_k))$}
            \myState {update the model parameters $\theta = \theta - \beta*\sum\limits_{k=1}\limits^{\mathcal{K}}g_k$}

            
        \EndFor
        \State
        \Require
        \State {initialize the homomorphic encryption method $E$}
        \State{build a task set $\mathcal{T}_c$ with size of $n_c$}
        \State{download the model parameters $\theta$ from server}
        \For{task $t_{i}$, $i = 1, 2, ..., n_c$}
                    \State copy the parameters $\theta$ as $\theta^{'}_i$
                    \For{$step$ = 1, 2, ..., $s_1$} 
                    \myState {use data in the support set to calculate $\mathcal{L}_{spt}(f_{\theta^{'}_i})$}
                    \myState {update $\theta^{'}_i = \theta^{'}_i - \alpha\bigtriangledown_{\theta^{'}_i}\mathcal{L}_{spt}(f_{\theta^{'}_i})$}

                    \EndFor
        
                 \myState {use data in the query set to calculate $\mathcal{L}_{qry}(f_{\theta^{'}_i})$}
                \EndFor
                 \myState {calculate the gradient $g_k = \bigtriangledown_{\theta}\sum\limits_{i=1}\limits^{n_{c}}\mathcal{L}_{qry}(f_{\theta^{'}_i})$}
                 \myState {encrypt the gradient $g_k$}
                 \myState {return the encryption result $E(g_k)$ to server}

	\end{algorithmic}
\end{algorithm}



\section{Experiments}

\subsection{Data Set Description}
\label{dataset}

Our data set \cite{githublink} consists of a total of 21,014 sample data points collected from eight distinct phones. Each data point contains nine features values and one label value. The feature set is presented in Table \ref{FEA} and a concise introduction to these features is in the appendices.

Table \ref{data} presents details of the data set. The data collection involved a total of eight distinct phones, of which two phones were of the same type. In the table, we refer to these two phones as OPPO R9m and OPPO R9m\_2 to differentiate between them. Additionally, we collected data using the OPPO R9m phone on two separate occasions, with a time span of 6 months between them. Consequently, the training data set contains two records with the same phone model ($C_1$ and $C_3$). In the subsequent experiments, we treat these data instances as if they were from two separate phones. The table also includes information about the operating system and the size of data set for each phone.

We designated six phones as contributors, while the remaining three phones are regarded as participants. As illustrated in Fig. \ref{dataset_fig}, the data from each phone was divided into two parts: 70\% of the data was allocated to the training data set, and the remaining 30\% was assigned to the validation data set. The combined training data set of contributors, denoted as $\mathcal{D}_{ct}$, comprised a total of 10,282 sample data points, while the combined validation data set $\mathcal{D}_{cv}$ consisted of 4,405 samples. Similarly, for the data set of participants, the sizes of $\mathcal{D}_{pt}$ and $\mathcal{D}_{pv}$ are 4,429 and 1,898, respectively.

\subsection{Crowdsourcing Group Construction}
\label{CGC}

Given that the majority of modules in our system operate within crowdsourcing groups, we will now elucidate the process of constructing the training group set and the validation group set utilizing the data from all 6 contributors:

\begin{itemize}
  \item \textbf{Step 1}: Randomly select $k$ phones from the pool of all 6 contributors, where $k$ is a randomly chosen number ranging from 2 to 6.
  \item \textbf{Step 2}: Generate the common label set of these $k$ phones, randomly pick one common label $T_{com}$ from the common label set.
  \item \textbf{Step 3}: For each selected phone, randomly choose one data sample that shares the same label as $T_{com}$.
\end{itemize}

Subsequently, the selected $k$ data samples will be considered to belong to the same crowdsourcing group. Following this criterion, we use the training data set $\mathcal{D}_{ct}$ to establish a training group set which contains 6000 crowdsourcing groups. Similarly, data from the validation data set $\mathcal{D}_{cv}$ is used to create a validation group set which contains 1500 groups.



\begin{table*}[]
\center
\setlength{\tabcolsep}{6mm}
\caption{The Description of data set.}
\label{data}
\begin{tabular}{llllll}
\hline
&ID& Phone Model& OS & Samp. & Total Samp.\\ \hline
\multirow{6}{*}{Contributors}&$C_1$ &OPPO R9m & Android 5.1 & 3139 & \multirow{6}{*}{14687}\\ 
&$C_2$& HUAWEI Nova7 Pro & HarmonyOS 2.0.0 & 1938 & \\  
&$C_3$& OPPO R9m & Android 5.1 & 2976 & \\ 
&$C_4$& HUAWEI SLA-AL00 & Android 7.0 & 2793 & \\ 
&$C_5$& OPPO R9sk & Android 6.0.1 & 2308 & \\ 
&$C_6$& HUAWEI ALP-AL00 & Android 10 & 1533 & \\ 
\hline

\multirow{4}{*}{Participants}&$P_1$& OPPO R11st & Android 8.1.0 & 1086 &\multirow{4}{*}{6327}\\ 
&$P_2$&MI 8 & Android 8.1.0 & 2274 & \\ 
&$P_3$& OPPO R9m\_2 & Android 5.1 & 2967 & \\\hline 

\end{tabular}
\center
\end{table*}

\subsection{Distributed Ambient Temperature Measurement Evaluation}

\subsubsection{Ambient Temperature Estimation Model Evaluation}
\label{section_temp_model}

\begin{figure*}
  \centering
  \subfloat[$C_1$: MAE = 0.277$^{\circ}$C]{\includegraphics[width=2.3in]{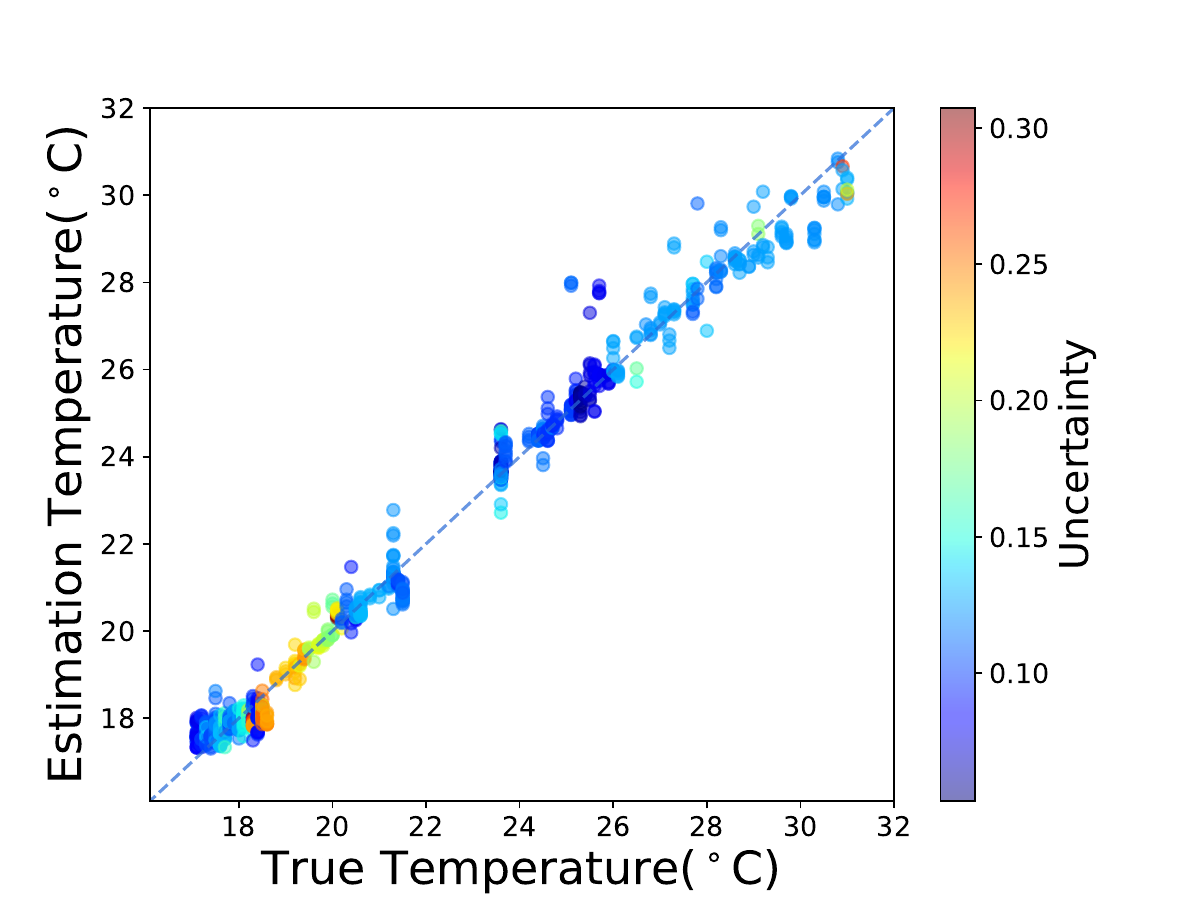}}
  \hfil
  \subfloat[$C_2$: MAE = 0.497$^{\circ}$C]{\includegraphics[width=2.3in]{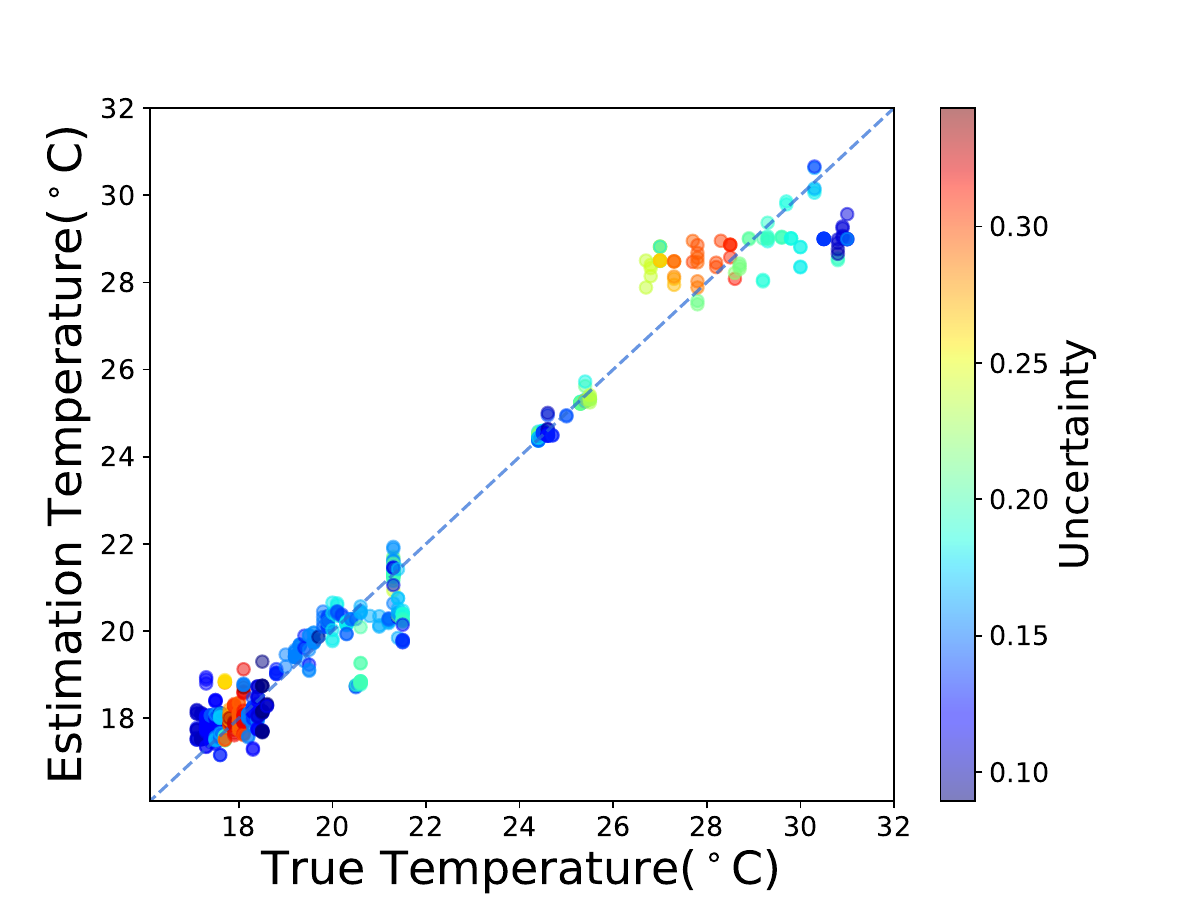}}
  \hfil
  \subfloat[$C_3$: MAE = 0.137$^{\circ}$C]{\includegraphics[width=2.3in]{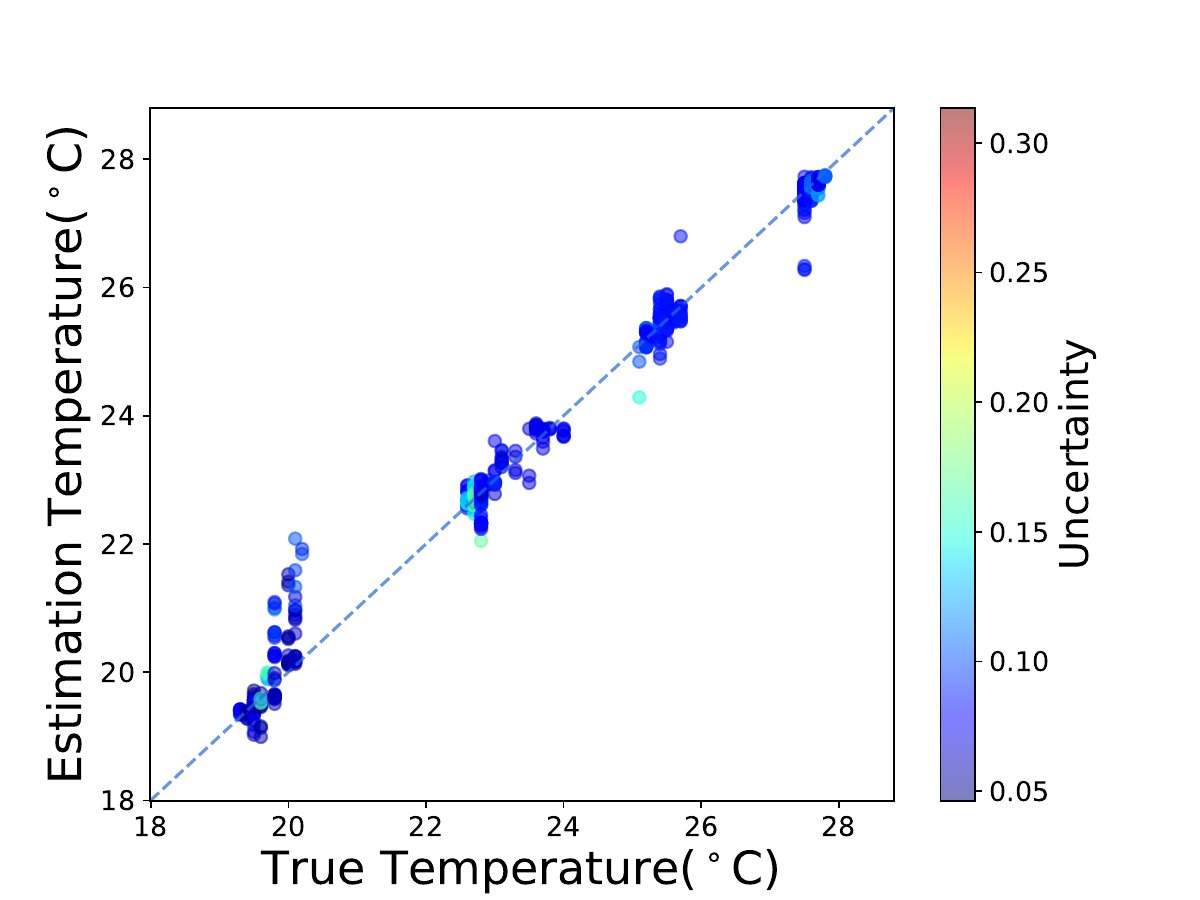}}\\

  \subfloat[$C_4$: MAE = 0.277$^{\circ}$C]{\includegraphics[width=2.3in]{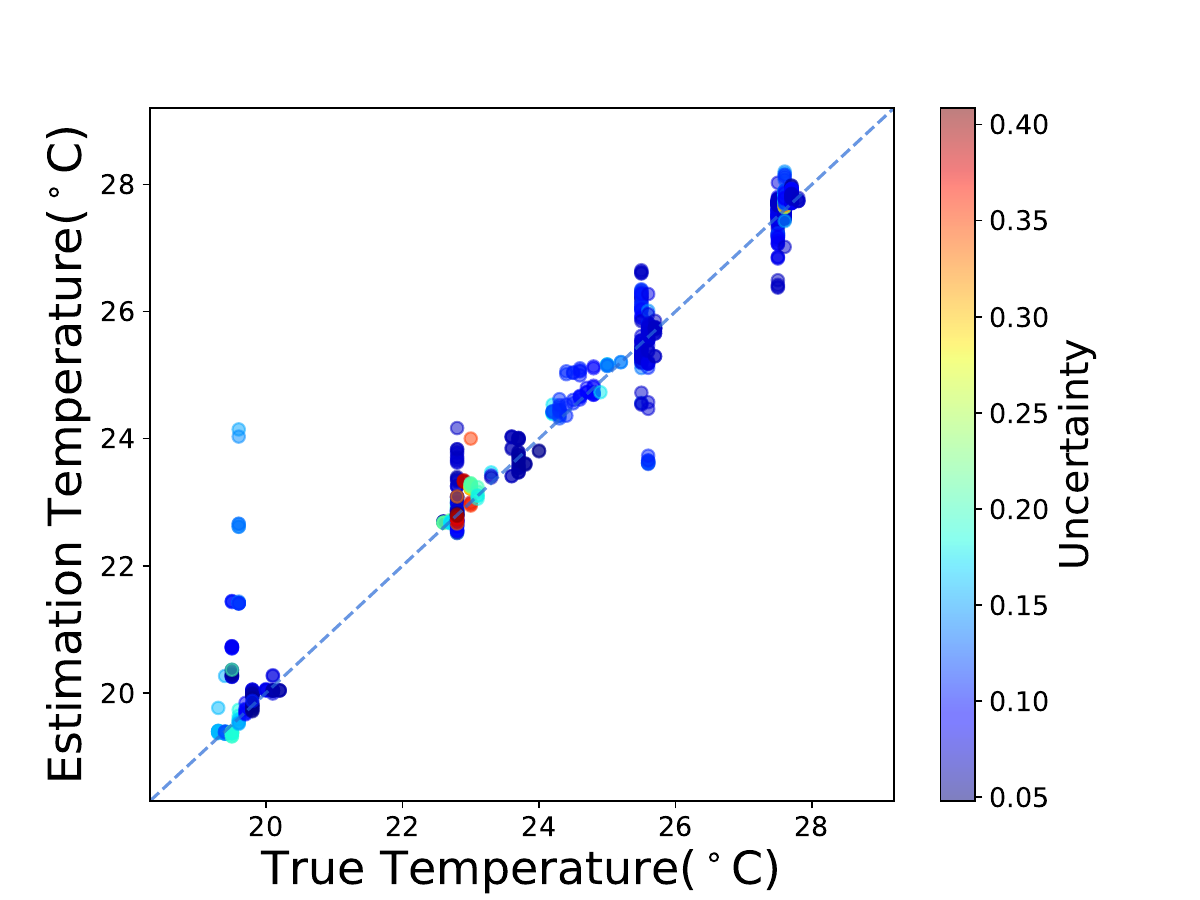}}
  \hfil
  \subfloat[$C_5$: MAE = 0.277$^{\circ}$C]{\includegraphics[width=2.3in]{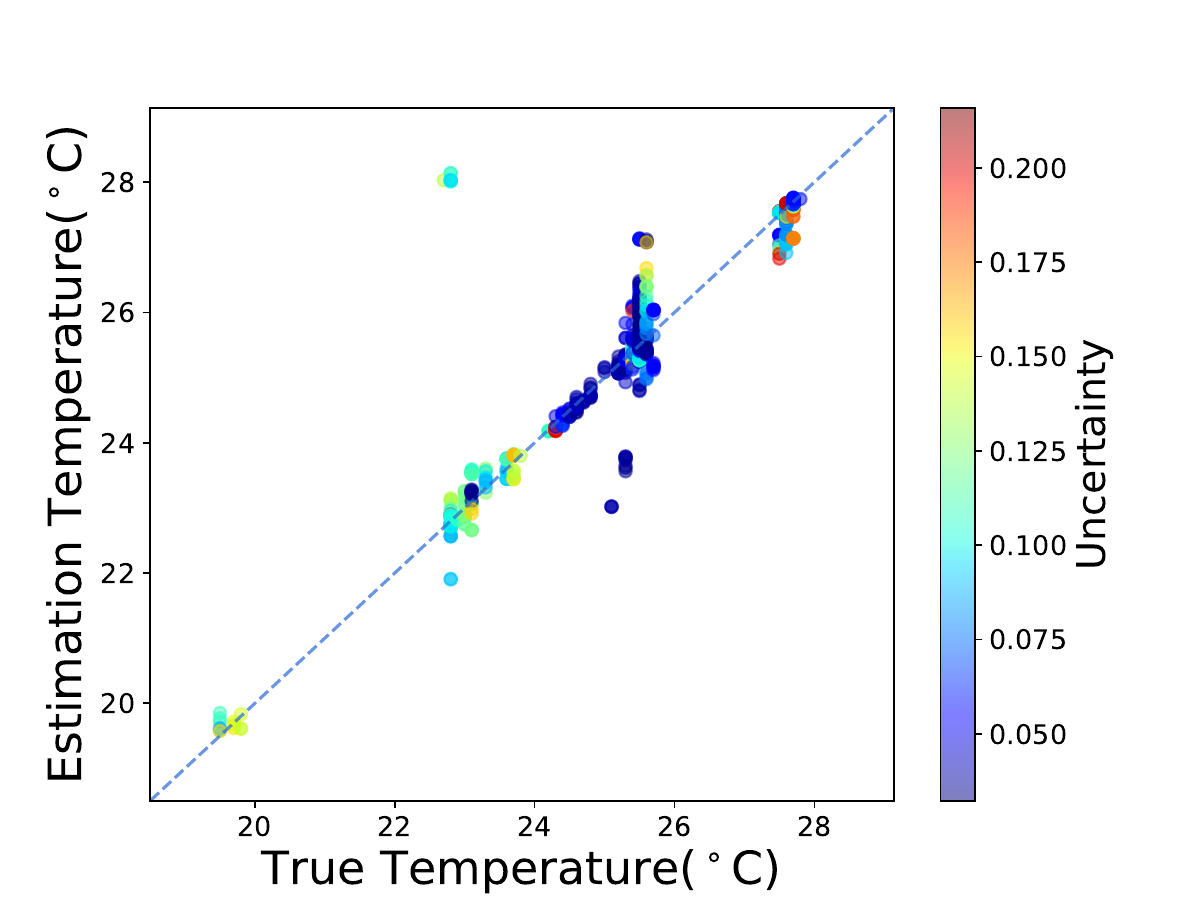}}
  \hfil
  \subfloat[$C_6$: MAE = 0.190$^{\circ}$C]{\includegraphics[width=2.3in]{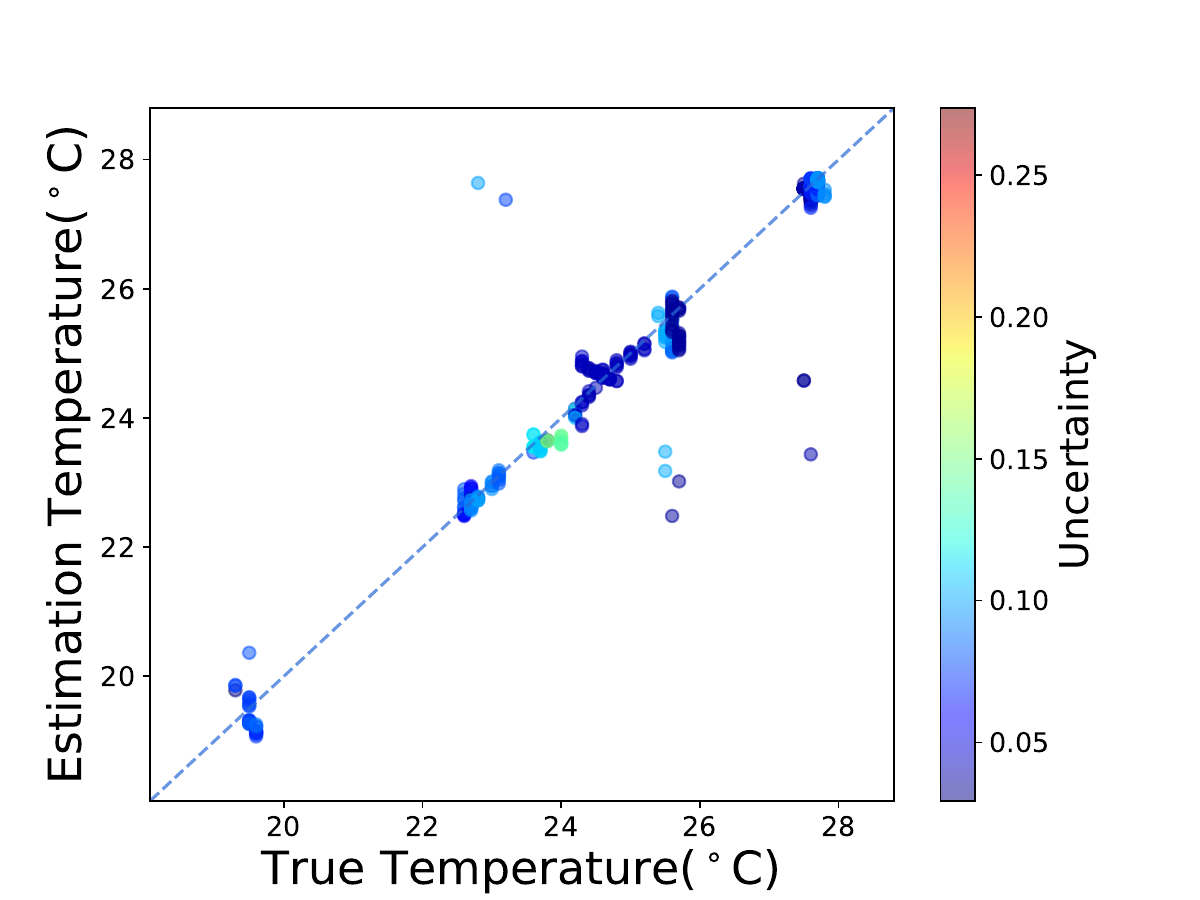}}
    \caption{
  The estimation results of all six contributors.}
  \label{single_prediction_diag}
\end{figure*}

In this section, we evaluate the performance of the ambient temperature estimation model on each single phone. The training data set of each phone is utilized to train a specific ambient temperature estimation model. During the training process, 20\% of the training data is reserved for determining the appropriate stopping point for training. Upon completion of the training process, the model of each contributor is evaluated on its respective validation data set.

Fig. \ref{single_prediction_diag} illustrates the evaluation results from all six contributors, showcasing the relatively accurate estimations achieved by each phone. Among them, phone $C_6$ demonstrates the best performance with a mean absolute error (MAE) of 0.190$^{\circ}$C. The highest MAE of 0.497$^{\circ}$C is observed on phone $C_2$. On average, the MAE across all contributors is 0.276$^{\circ}$C. We leave the more comprehensive analysis of the estimation model in the appendices.


\subsubsection{CBTS Truth Inference Model Evaluation}


We use the training group set described in section \ref{CGC} to train the CBTS model until the loss on the validation group set ceased to decrease for a consecutive 20 epochs. During training, the gradients are back-propagated only once all $k$ data instances have been aggregated into a final result. Subsequently, we utilized the data set $\mathcal{D}_{cv}$ to reconstruct a testing group set consisting of 6000 distinct groups to evaluate the performance of the CBTS model.

Table \ref{cs_baseline} presents the evaluation results of the CBTS model and seven other baseline methods. The first three methods, namely D\&S \cite{DS}, PM \cite{PM}, and ZC \cite{ZC}, are typical truth inference algorithms that are applicable to numeric tasks. The next two methods, MV-2 and MV-3, are modified versions of the Majority Voting (MV) \mbox{\cite{cs_all}} approach. In these two methods, we clustered all the answers and selected the average of the largest cluster as the final answer. The number associated with each method indicates the set cluster number. The final two methods are the aggregation approaches utilized in other similar phone-based ambient temperature measurement studies. The Mean method, employed in works \cite{2,3,4}, calculates the average of all the given answers to derive the final result. On the other hand, the Weighted Average (WA) method, utilized in the work \cite{4}, assigns a confidence value to each phone based on its historical performance and employs a weighted average approach using these confidence values. For a more comprehensive understanding of these baseline methods and the rationale behind the baseline selection, we provide a detailed description in the appendices.

The MAE of the crowdsourcing results derived from the CBTS model is 0.136$^\circ$C. Compared to the ambient temperature measurement by a single phone, which has an average error of 0.276$^{\circ}$C, the application of crowdsourcing techniques reduces the error by approximately 50\%. From the results in Table \ref{cs_baseline}, we see the CBTS model surpasses the performance of all other methods listed. Additionally, we assessed each method under varying group scales. The findings indicate that, in comparison to other algorithms, the CBTS model and the revised MV method exhibit less sensitivity to group scales. Remarkable results can be achieved with the CBTS model even when the group size is as small as 3. Although the performance of MV could be very close to CBTS, CBTS is much more flexible than the MV method. Because for MV method, we are required to set clustering parameters in advance based on the number of participants. Moreover, the results of MV can be obtained only when collecting enough answers. However, the CBTS model can provide a result no matter how many answers collected.



\begin{table*}[]
\center
\setlength{\tabcolsep}{3mm}
\caption{
The MAEs of the CBTS model and other baseline truth inference algorithms which are evaluated on the testing group set across different group scales.
}
\label{cs_baseline}
\begin{tabular}{lllllllll}
\hline
&CBTS&  D\&S\cite{DS} & PM\cite{PM} & ZC\cite{ZC} & \mbox{MV-2\cite{cs_all}} & \mbox{MV-3\cite{cs_all}} & \mbox{Mean\cite{cs_all}} & \mbox{WA\cite{4}}\\ \hline
2 Phones&\textbf{0.205}&  0.504 & 0.266 & 0.294& 0.232& 0.232 & 0.232& 0.220\\  
3 Phones& 0.139&  0.286 & 0.180 & 0.252& \textbf{0.125}& 0.185 & 0.185& 0.173\\ 
4 Phones&\textbf{0.119}&  0.281 & 0.146 & 0.198& 0.137& 0.123 & 0.179& 0.159\\ 
5 Phones& 0.110&  0.273 & 0.122 & 0.158& \textbf{0.106}& 0.127 & 0.166&0.141\\ 
6 Phones&0.109&  0.314 & \textbf{0.089} & 0.193& 0.109& 0.103 & 0.164&0.131\\ 
All Groups&\textbf{0.136}& 0.333 & 0.160 & 0.218& 0.141& 0.154 & 0.185 & 0.164\\ 
\hline

\end{tabular}
\center
\end{table*}

\subsection{Few-shot Learning with The Inferred Label}

\subsubsection{Automatic Inferred Label Generation}

To evaluate the effectiveness of annotating new data using the CBTS truth inference model, we conduct the inferred label generating experiments on the training data set of participants, denoted as $\mathcal{D}_{pt}$. 

Following the same group construction rules as before, for each data instance $x_i$ in $\mathcal{D}_{pt}$, we randomly sample $k$ data instances from k different contributors in the data set $\mathcal{D}_{c}$. All of these data instances are required to have the same label as $x_i$, so that we can assume they are collected from the same crowdsourcing group at the same time. By taking this rule, every crowdsourcing group contains $1$ data instance from $1$ participant and $k$ data instances from $k$ different

For each crowdsourcing group, we do the following operations:
\begin{itemize}
  \item \textbf{Step 1}: Compute the estimation results of all $k$ contributor data instances using their estimation models.
  \item \textbf{Step 2}: Take the CBTS model to crowdsource all of these results, obtaining a final inferred answer.
  \item \textbf{Step 3}: Assign the inferred answer to the remaining data instance $x_i$, designating it as the inferred label.
\end{itemize}

\begin{figure*}
  \centering
 \subfloat[$P_1$: MAE = 0.138$^{\circ}$C]{\includegraphics[width=2.3in]{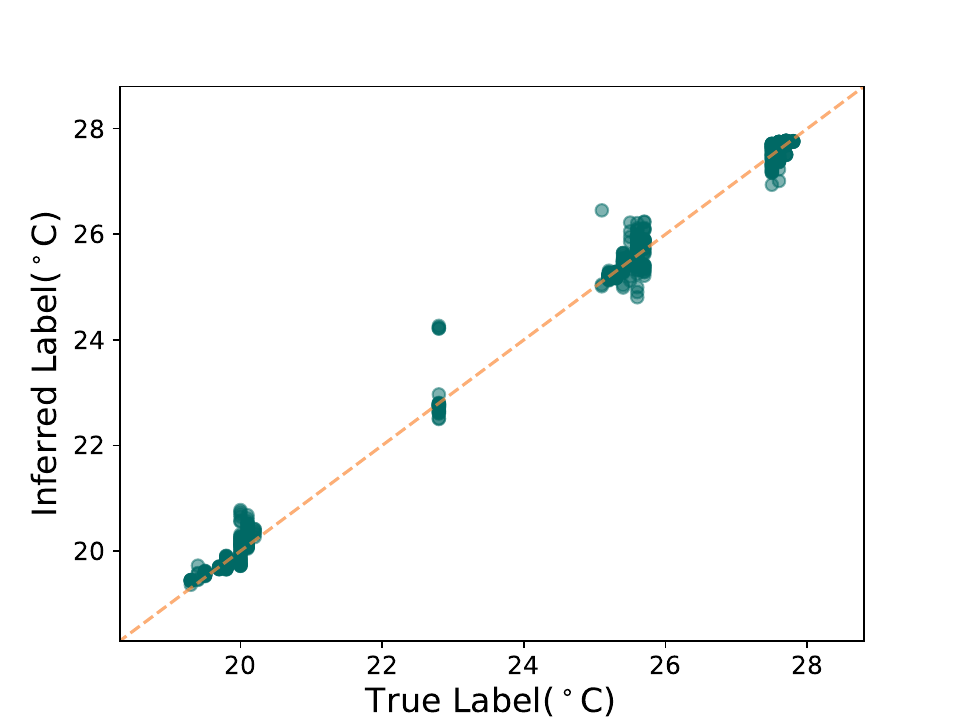}}
  \hfil
  \subfloat[$P_2$: MAE = 0.168$^{\circ}$C]{\includegraphics[width=2.3in]{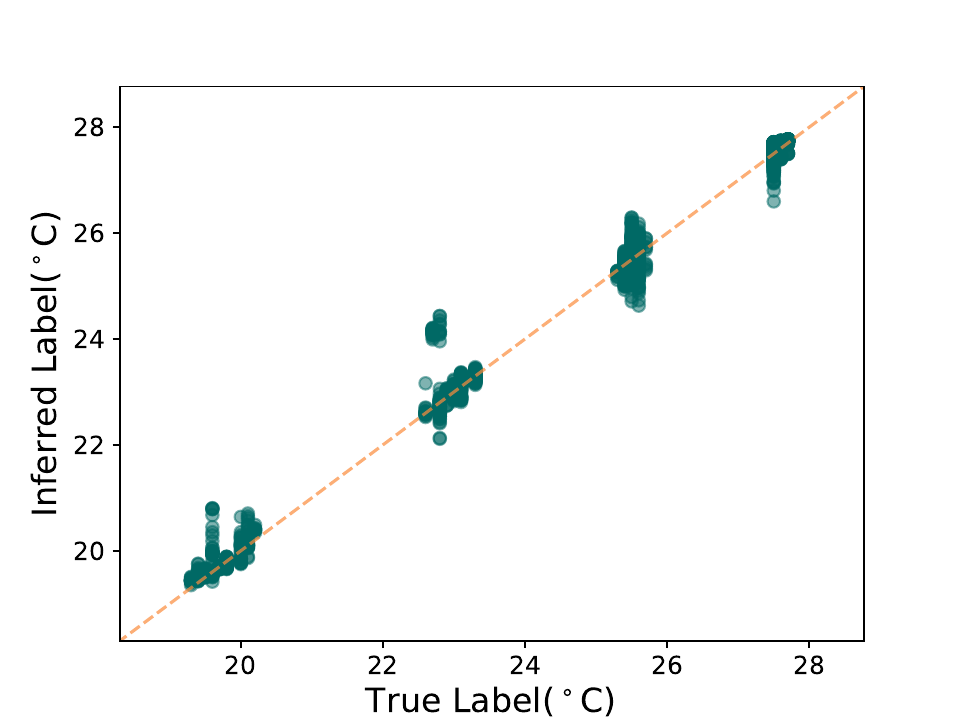}}
  \hfil
  \subfloat[$P_3$: MAE = 0.177$^{\circ}$C]{\includegraphics[width=2.3in]{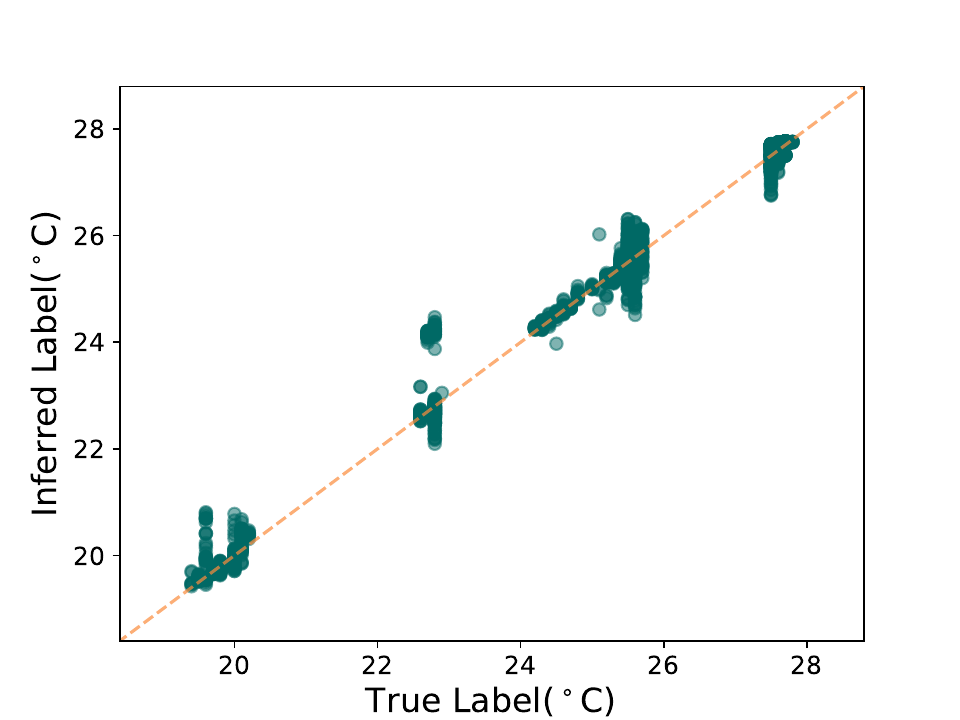}}
  \caption{The comparison plots of the inferred label and the true label on the whole training data set of all participants ($P_1$ to $P_3$). In every sub-figure, the horizontal axis represents the true label, while the vertical axis portrays the inferred label.}
  \label{infer_label}
\end{figure*}

We generate the inferred label for each data instance $x_i$ in $\mathcal{D}_{pt}$ and compare them with the true labels. Fig. \ref{infer_label} illustrates the bias between the true labels and the inferred labels. Moreover, we provide the specific errors in each sub-figure. From the results, we can see the average error between inferred labels and true labels on all 3 phones is only 0.161$^{\circ}$C, which is very small.

\subsubsection{Few-shot Learning with MAML Framework}

In this section, we explore various few-shot learning strategies to train a new temperature estimation model for a newly added participant who possesses only 5 data instances. To further evaluate the efficacy of inferred labels, we train each model using both true labels and inferred labels. Apart from the MAML training strategy, we consider several baseline training methods, which are as follows:
\begin{itemize}
  \item \textbf{Pre-training (PT)}: In this method, we employ the entire training data set of all contributors, denoted as $\mathcal{D}_{ct}$, to pre-train a model until convergence. Subsequently, we fine-tune the model using the 5 data instances collected from the selected participant.
  \item \textbf{Direct-training (DT)}: This method involves training a model from scratch using the 5 data instances collected from the chosen participant.
\end{itemize}

\begin{table*}[]
\center
\setlength{\tabcolsep}{6mm}
\caption{The performance of models that are trained with different method by using 5 pieces of data on each participant for a 20-step update. We measure them with MAE($^{\circ}$C).}
\label{res_all_participants}
\begin{tabular}{lllllll}
\hline
 Participant&  DT-TL  & DT-IL & PT-TL &  PT-IL&  MAML-TL &  MAML-IL \\ \hline
$P_1$ & 2.613 &2.609 &2.253 & 2.255 &0.834 & \textbf{0.814} \\
$P_2$& 2.391 &2.401 &1.430 & 1.433 &\textbf{1.355} & 1.402
\\
$P_3$& 2.423 &2.431 & 1.412 & 1.427 &\textbf{0.829} & 0.852 \\
$All$& 2.444 & 2.451 & 1.563 & 1.571 & \textbf{1.019} & 1.043 \\
\hline
\end{tabular}
\center
\end{table*}

\begin{figure*}

  \centering
 \subfloat[$P_1$, Direct-training]{\includegraphics[width=2in]{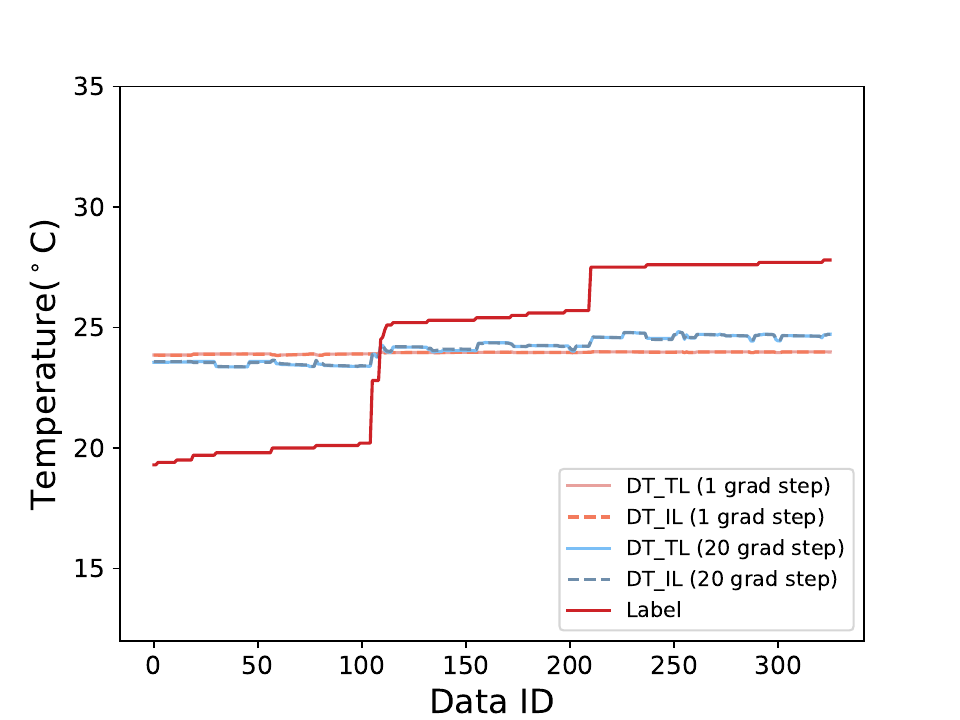}}
  \hfil
  \subfloat[$P_1$, Pre-training]{\includegraphics[width=2in]{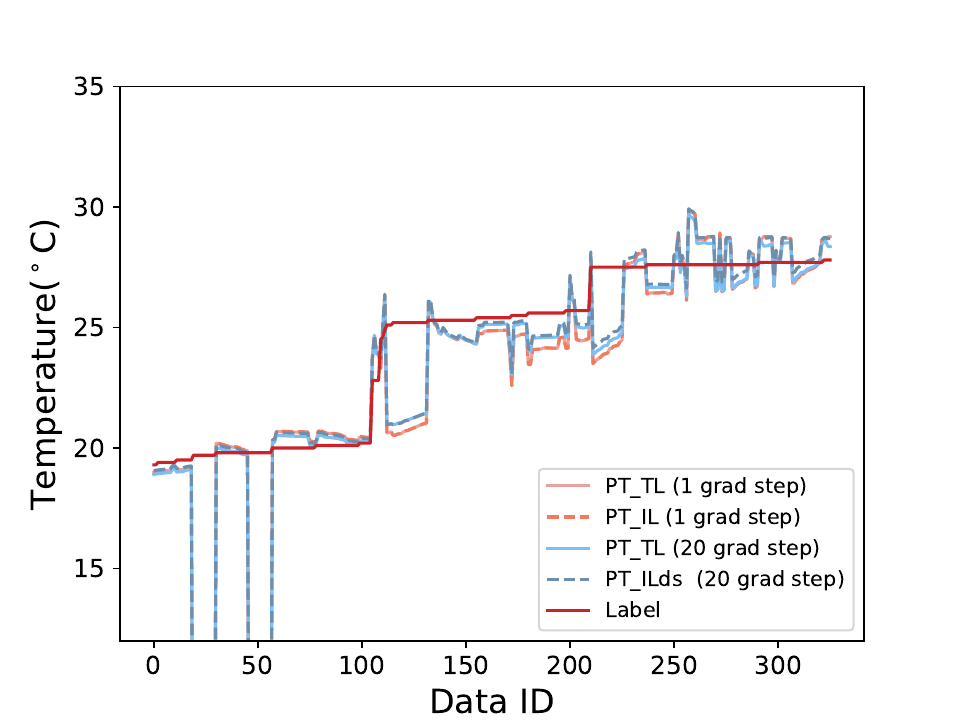}}
  \hfil
  \subfloat[$P_1$, MAML]{\includegraphics[width=2in]{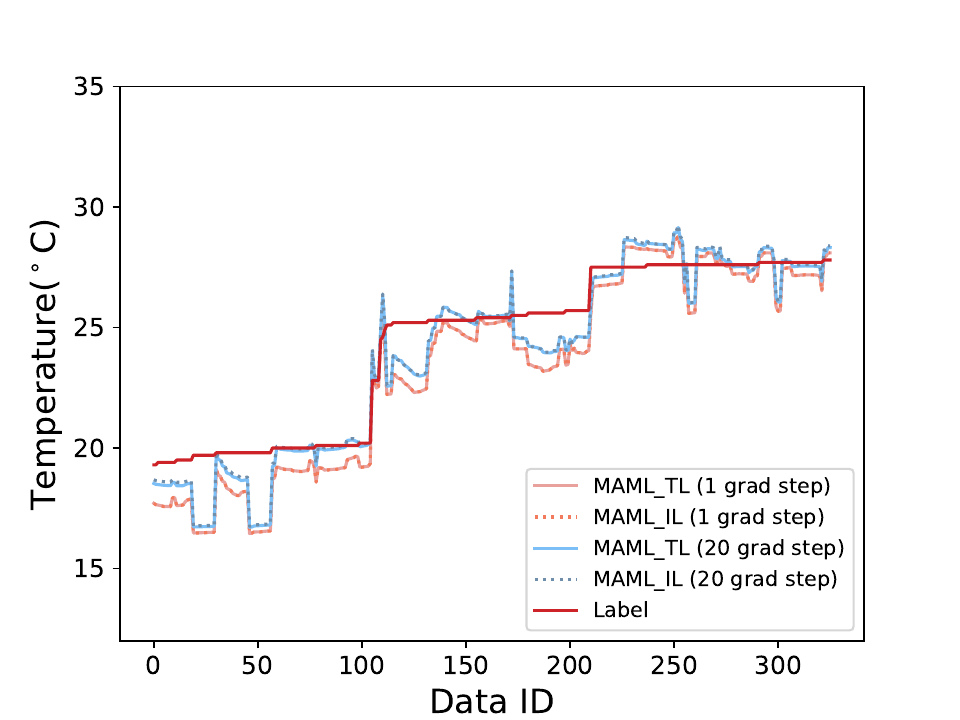}}\\
 \subfloat[$P_2$, Direct-training]{\includegraphics[width=2in]{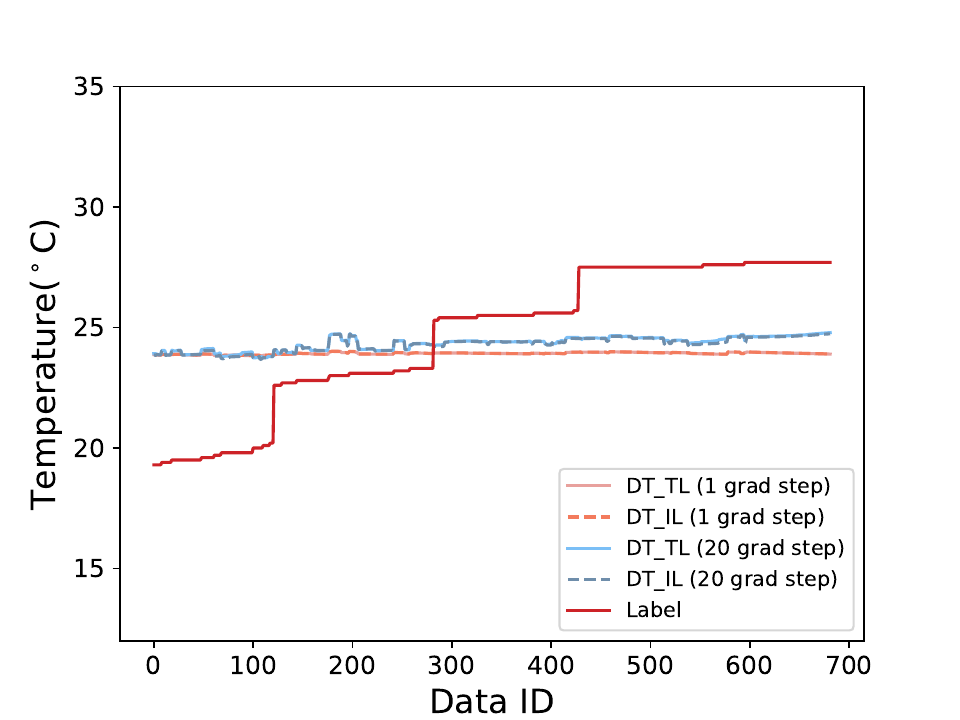}}
  \hfil
  \subfloat[$P_2$, Pre-training]{\includegraphics[width=2in]{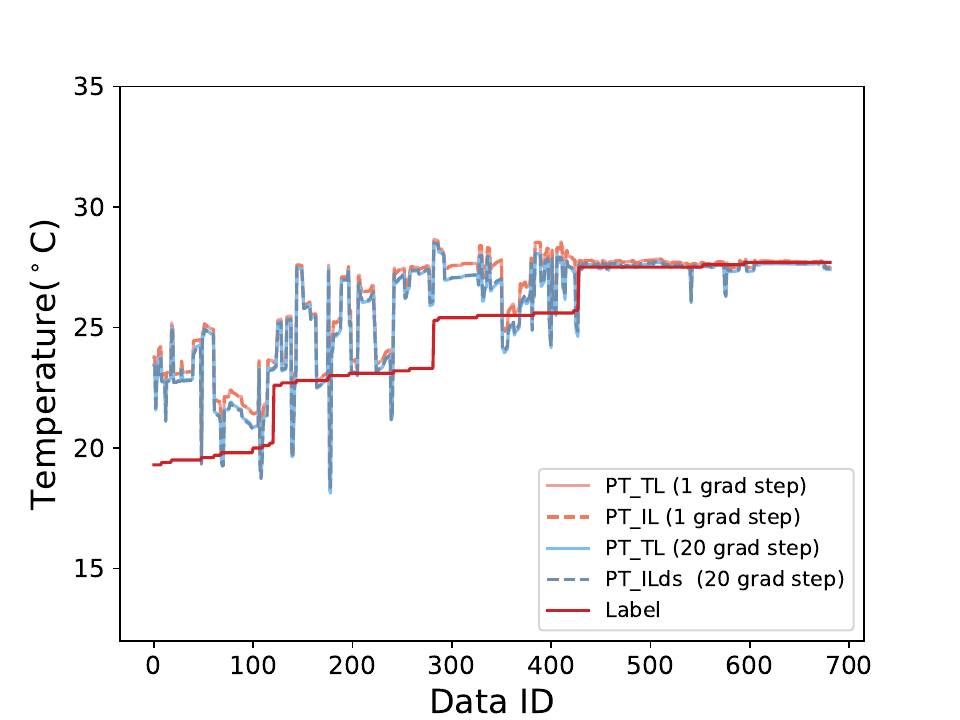}}
  \hfil
  \subfloat[$P_2$, MAML]{\includegraphics[width=2in]{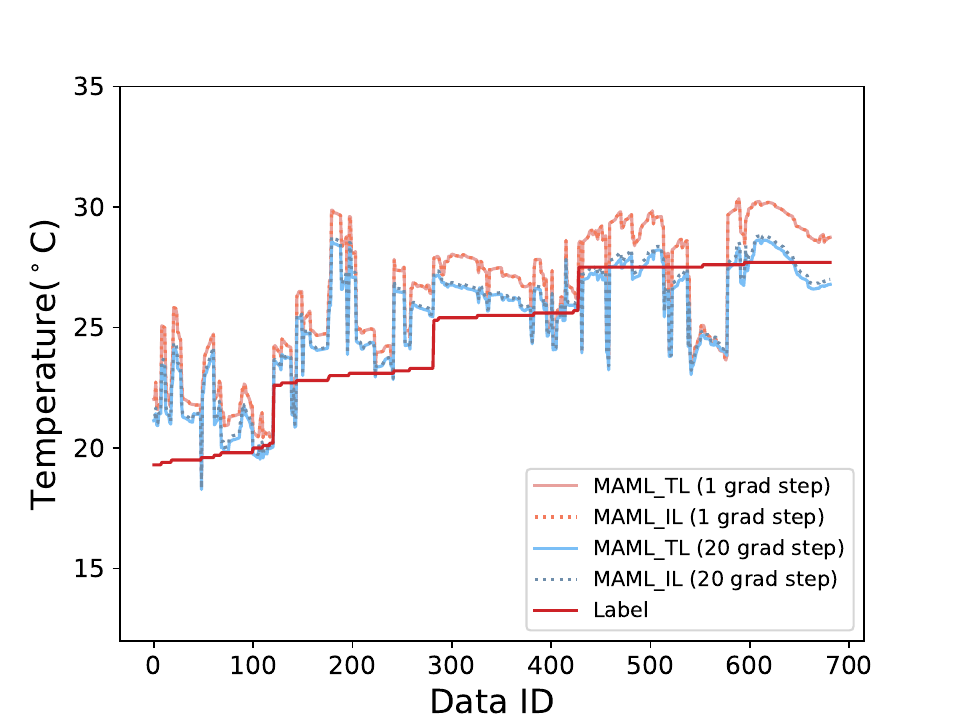}}\\
   \subfloat[$P_3$, Direct-training]{\includegraphics[width=2in]{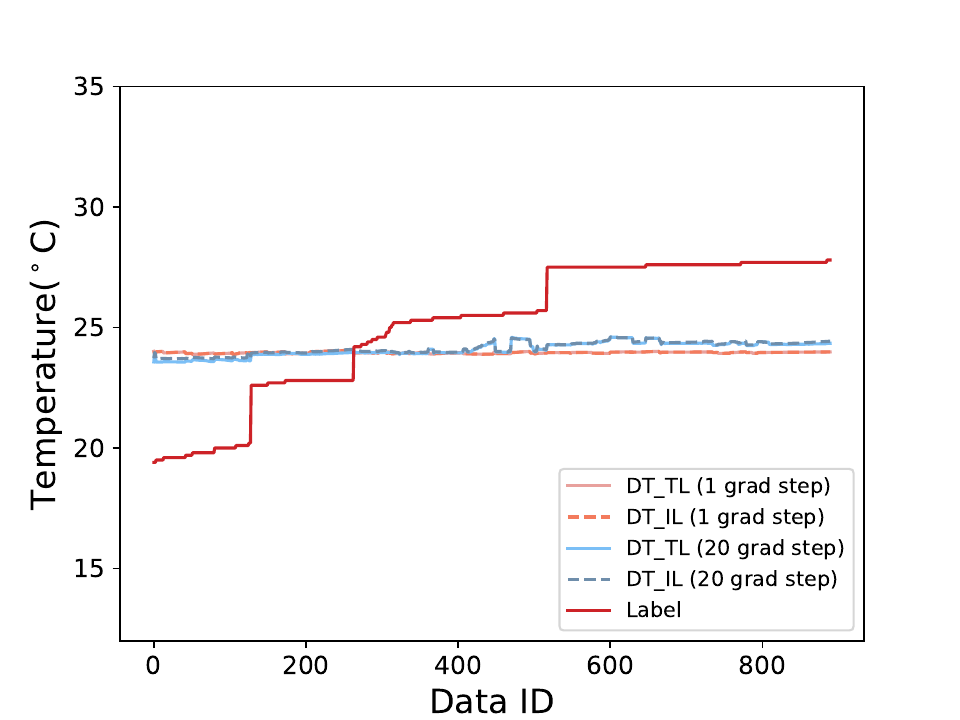}}
  \hfil
  \subfloat[$P_3$, Pre-training]{\includegraphics[width=2in]{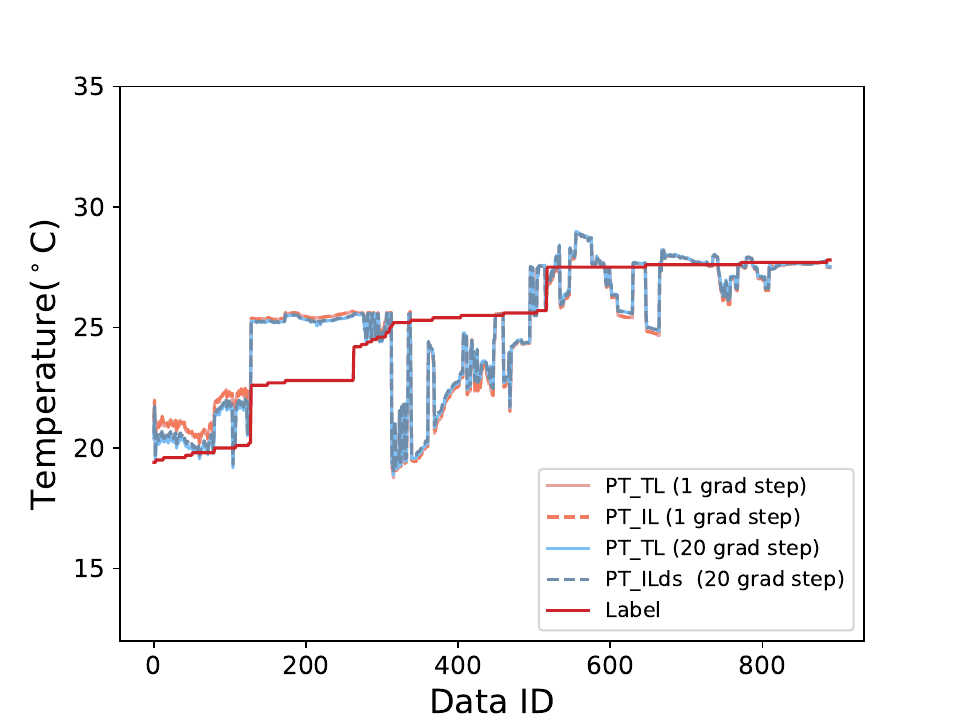}}
  \hfil
  \subfloat[$P_3$, MAML]{\includegraphics[width=2in]{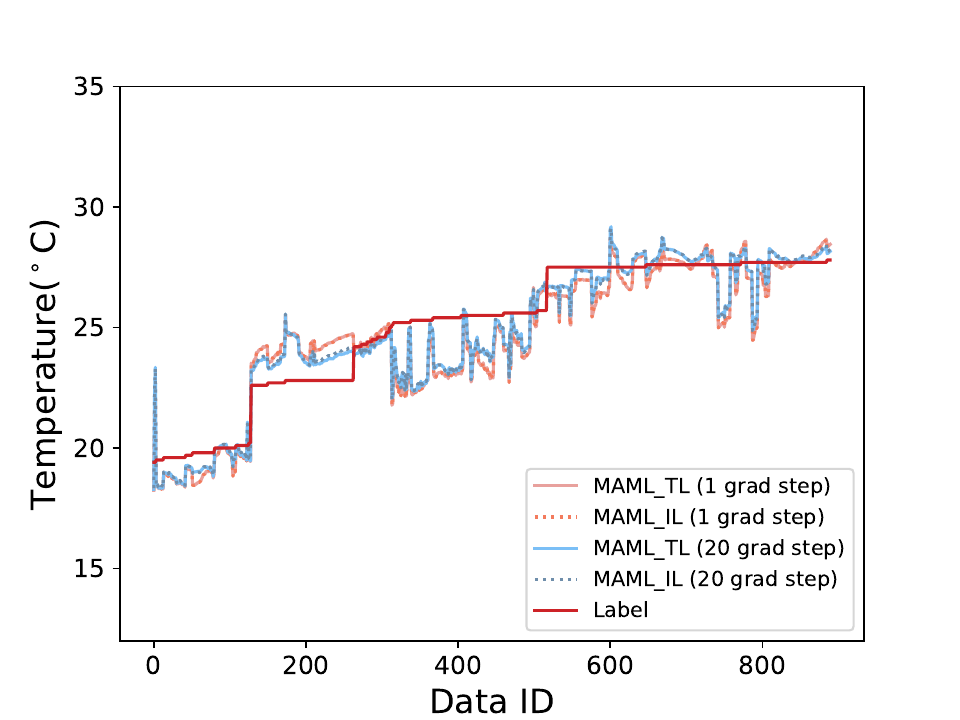}}\\

  \caption{The figure depicts the results of few-shot learning using the Direct-training (DT), Pre-training (PT), and Model-Agnostic Meta-Learning (MAML) methods. In order to maintain visual appeal, the y-axis range for all the images has been set between 12 and 35. Consequently, some of the outliers in figure (b) are not visible due to the restricted y-axis range.}
  \label{dt_pt_maml}
\end{figure*}

To better evaluate each strategy, we repeat the process of training and evaluation for 100 times. We use each strategy to train 100 models and evaluate their performance on the complete validation data set $\mathcal{D}_{cvn}$ of each participant $P_n$. For each training iteration of a specific participant, we randomly select 5 data instances from their corresponding training data set $\mathcal{D}_{ctn}$. We also fine-tune each model for both 1 step and 20 steps, evaluating their performance accordingly. The results are illustrated in Fig. \ref{dt_pt_maml} and summarized in Table \ref{res_all_participants}. In all experiments conducted, adhering to the notation settings in Algorithm \mbox{\ref{algo1} and Algorithm \ref{algo2}}, we set the task-level learning rate ($\alpha$) to 0.001 and the meta-learning rate ($\beta$) to 0.01. We chose a task batch size ($n$) of 400, with the task-level update steps ($s_1$) being 5. Depending on the specific experiment, the number of meta-level update steps could be either 1 or 20. For optimization, we employed the Adam optimizer.

 In Table \ref{res_all_participants} and Fig \ref{dt_pt_maml}, ``TL'' indicates the true label and ``IL'' represents the inferred label. The final row of Table \mbox{\ref{res_all_participants}} presents the model performance across the evaluation data sets from all three participants. Upon comparing the plots in each figure, we can observe that utilizing inferred labels to train a model does not introduce significant deviations in the results. When all other conditions remain constant, the prediction curves obtained by training with true labels closely align with those obtained by training with inferred labels. Interestingly, for participant $P_1$, some methods achieve superior performance with inferred labels compared to true labels. This could be attributed to the relatively small data quantity for $P_1$, which may introduce a degree of randomness. However, when examining the results across all participants, it is evident that the overall performance when utilizing inferred labels is marginally lower than that achieved with true labels.

Among the various training strategies, MAML exhibits a substantial advantage over others. Taking into account the average performance across three participants, MAML surpasses Direct-learning by 59\% and exceeds Pre-training by 36\%. The results presented in the tables unequivocally demonstrate that training with MAML consistently achieves the lowest MAE across all three participants. These results indicate that even with only 5 new data instances, by training with MAML, we can effectively reduce the estimation error to less than 1$^{\circ}$C. Furthermore, we assessed the computational efficiency by fine-tuning over 20 epochs on our CPU (12th Gen Intel(R) Core(TM) i7-12650H 2.30 GHz), where the average duration across 100 trials was merely 0.017 seconds. Given that our data collection app samples data at a rate of 5 seconds, this allows for the rapid deployment of a new estimation model onto a new phone. This is of great significance for the extension of the temperature measurement model in practice.

\subsubsection{Federated Learning}

As illustrated in the methodology section, the incorporation of federated learning into the system does not alter the original process of few-shot learning. We use the TenSEAL library \mbox{\cite{tenseal2021}} to evaluate the performance of applying CKKS on the gradients. During the experiment, we set 100 virtual clients participate in cryptographic gradient computation for 100 times and every time we randomly generate a gradient for each client. The average time of encryption, summation and decryption on 100 clients is only 7.306 ms. The average error with respect to the result calculated in plain text is only 1.335e-10. The result is almost exactly the same as those obtained by direct calculation.

\section{Future Work}

This study presents a comprehensive technical overview of the system implementation. In future work, we will focus on enhancing the user experience and usability of the system. Specifically, we may try to establish a reward mechanism based on the duration of participation. Then we will also address and discourage the related malicious competition behavior. Moreover, We will try to integrate personal phone-based temperature estimation with various appliances and electronic devices to incentivize continuous participation.



\section{Conclusion}

In this study, we present a phone-based distributed ambient temperature measurement system with an efficient label-free automated training strategy. The primary objective of our system is to enable collaboration between multiple smartphones in large indoor spaces for real-time measurement of ambient temperature in different small areas. Through crowdsourcing, the system can achieve an MAE of only 0.136$^{\circ}$C in temperature measurement. Additionally, our system incorporates an automatic data annotation method that leverages crowdsourcing technology to provide labels for newly collected data. The MAE between the generated labels and the true labels is only 0.161$^{\circ}$C. To expedite the training of an accurate temperature estimation model for newly added phones, we employ a few-shot training framework, which achieves an MAE of 0.814$^{\circ}$C with just 5 data points. Furthermore, we demonstrate the integration of federated learning into our system, ensuring the protection of privacy. With this work, we aim to advance the practical application of phone-based ambient temperature estimation and pave the way for further energy-saving technologies. Our system has the potential to contribute to energy-efficient practices and encourage the adoption of innovative solutions in various domains.

\section*{Acknowledgments}
This work was supported by the RISUD Projects (No. P0042845 and No. 1-BBWW) and the RISE Project (No. P0051003) of The Hong Kong Polytechnic University, International Centre of Urban Energy Nexus of The Hong Kong Polytechnic University (No. P0047700), and KKS Synergy project, Energy flexibility through synergies of big data, novel technologies \& systems, and innovative markets (20200073).


 
%
\bibliographystyle{ieeetr}
\bibliography{sample-base}

\section{Biography Section}
 

\begin{IEEEbiography}[{\includegraphics[width=1in,height=1.25in,clip,keepaspectratio]{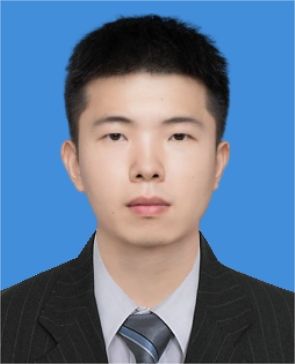}}]{Dayin Chen} is currently pursuing his Ph.D. in Building Energy and Environment Engineering at The Hong Kong Polytechnic University. He holds both a Bachelor's and a Master's degree in Computer Science from the Southern University of Science and Technology. His research interests include mobile computing, crowdsourcing, and neural architecture search. Additionally, Chen is exploring the application of computer science technologies in enhancing building energy efficiency.
\end{IEEEbiography}

\begin{IEEEbiography}[{\includegraphics[width=1in,height=1.25in,clip,keepaspectratio]{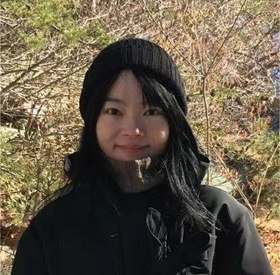}}]{Xiaodan Shi} received the B.E. and M.S. degrees in photogrammetry and remote sensing from Wuhan University, China. She  received the Ph.D. degree with the Center for Spatial Information Science, The University of Tokyo, Kashiwa, Japan. She is working as a researcher with the Center for Spatial Information Science, The University of Tokyo. Her current research interests include computer vision and its applications in pedestrian trajectory prediction, human flow monitoring, multi-objects tracking and data mining in GPS.
\end{IEEEbiography}

\begin{IEEEbiography}[{\includegraphics[width=1in,height=1.25in,clip,keepaspectratio]{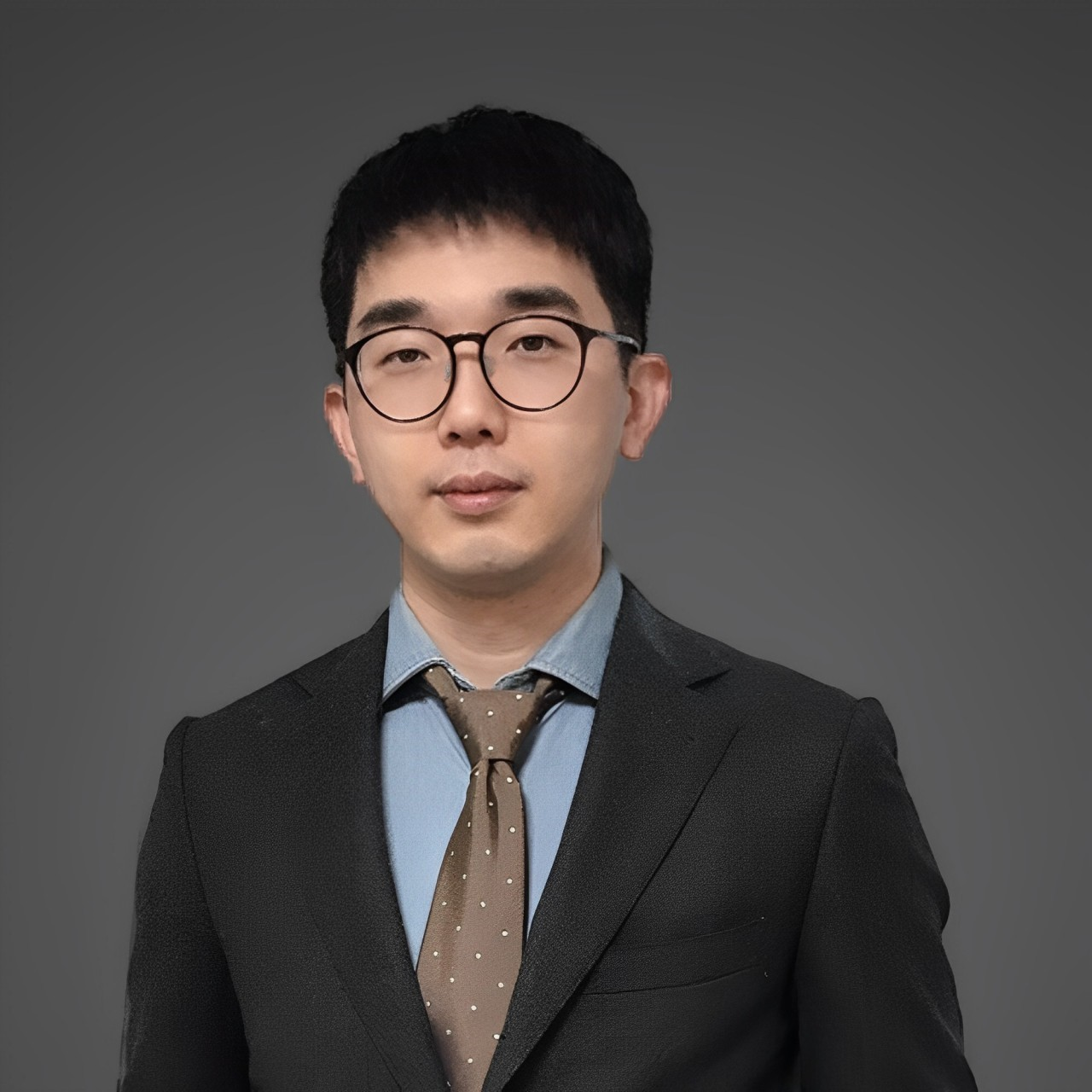}}]{Haoran Zhang} is a researcher with an interdisciplinary background, holding dual Bachelor's degrees in Engineering and Economics, complemented by two Ph.D. degrees in Oil \& Gas Storageand Transportation Engineering ond Environmental science. His research is at the forefront of sustainable development, focusing on clean energy supply chains, green transportation, and smart urban energy systems. His innovative research has garnered prestigious accolades, including the Energy Globe Award, the R\&D 100 Award, the John Tiratsoo Award for Young Achievement, and the Smart 50 Award among others.
\end{IEEEbiography}


\begin{IEEEbiography}[{\includegraphics[width=1in,height=1.25in,clip,keepaspectratio]{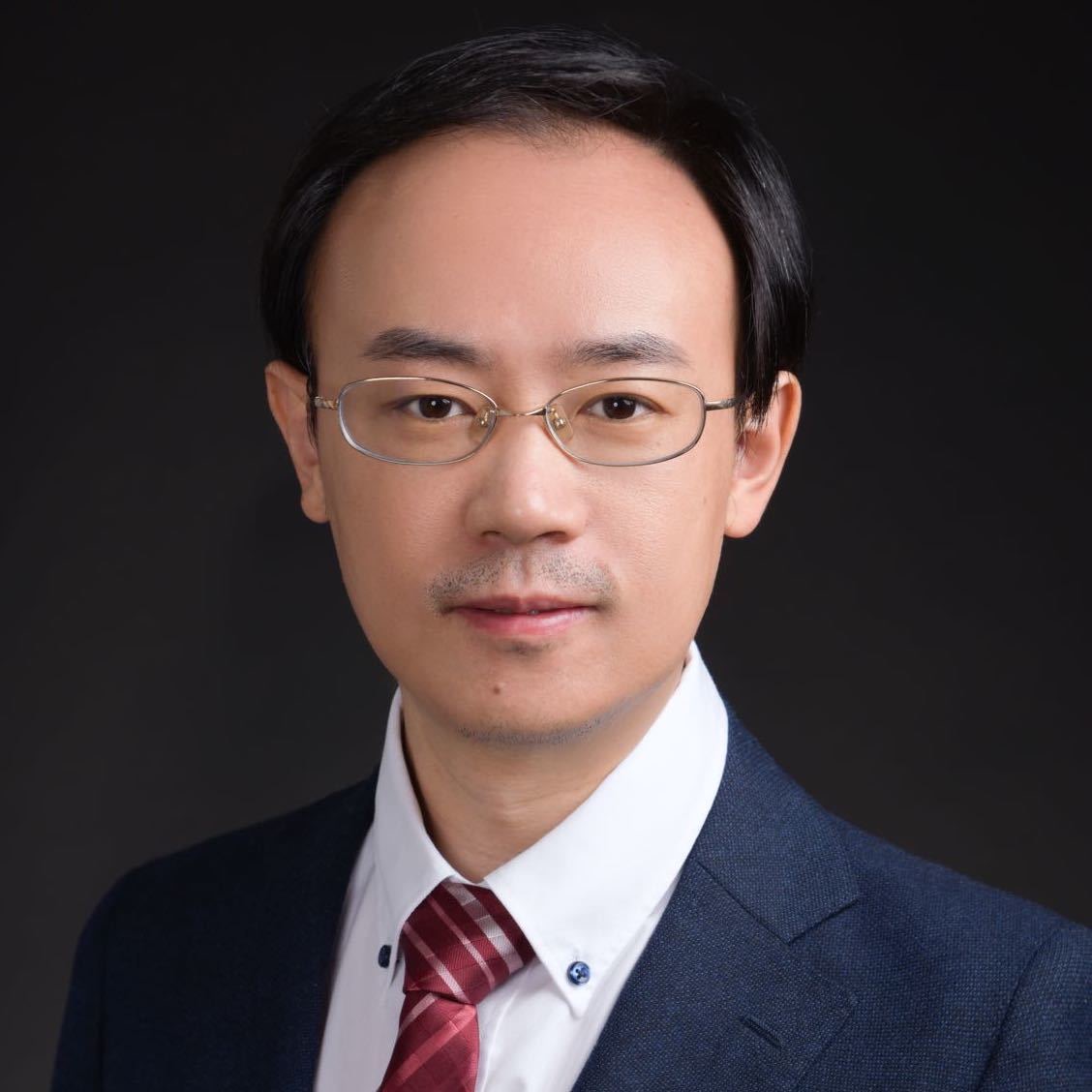}}]{Xuan Song} received the Ph.D. degree in signal and information processing from Peking University in 2010. In 2017, he was selected as Excellent Young Researcher of Japan MEXT. He served as Associate Editor, Guest Editor, Area Chair, Senior Program Committee Member for many famous journals and top-tier conferences, such as IMWUT, IEEE Transactions on Multimedia, WWW Journal, ACM TIST, IEEE TKDE, Big Data Journal, UbiComp, IJCAI, AAAI, ICCV, CVPR and etc. His main research interest are AI and its related research areas, such as data mining and urban computing. By now, he has published more than 160 technical publications in journals, book chapter, and international conference proceedings, including more than 110 high-impact papers in top-tier publications for computer science. His research was featured in many Chinese, Japanese and international media, including United Nations, the Discovery Channel, and Fast Company Magazine. He received Honorable Mention Award in UbiComp 2015.
\end{IEEEbiography}

\begin{IEEEbiography}[{\includegraphics[width=1in,height=1.25in,clip,keepaspectratio]{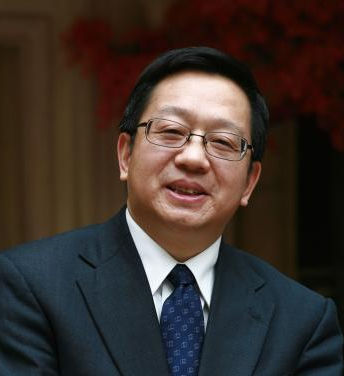}}]{Dongxiao Zhang} received the M.S. and Ph.D. degrees in hydrology from the University of Arizona, Tucson, AZ, USA, in 1992 and 1993, respectively. From 1996 to 2003, he was a Technical Staff Member and the Team Leader with the Los Alamos National Laboratory, Los Alamos, NM, USA. From 2004 to 2007, he was the Miller Chair Professor of petroleum and geological engineering with the University of Oklahoma, Norman, OK, USA. From 2007 to 2010, he was the Marshall Professor of the
Petroleum Engineering Program with the University of Southern California, Los Angeles, CA, USA. From 2010 to 2019, he was the Chair Professor and the Dean of the College of Engineering, Peking University, Beijing, China. He was the Provost and the Chair Professor with the Southern University of Science and Technology, Shenzhen, China. He is currently the Provost and the Chair Professor with the Eastern Institute of Technology, Ningbo, China. His research interests include stochastic uncertainty quantification and inverse modeling, mechanisms for shale–gas and coalbed–methane production, and geological carbon sequestration. Dr. Zhang is a member of the U.S. National Academy of Engineering, a fellow of the Geological Society of America, and an Honorary Member of the Society of Petroleum Engineers.
\end{IEEEbiography}

\begin{IEEEbiography}[{\includegraphics[width=1in,height=1.25in,clip,keepaspectratio]{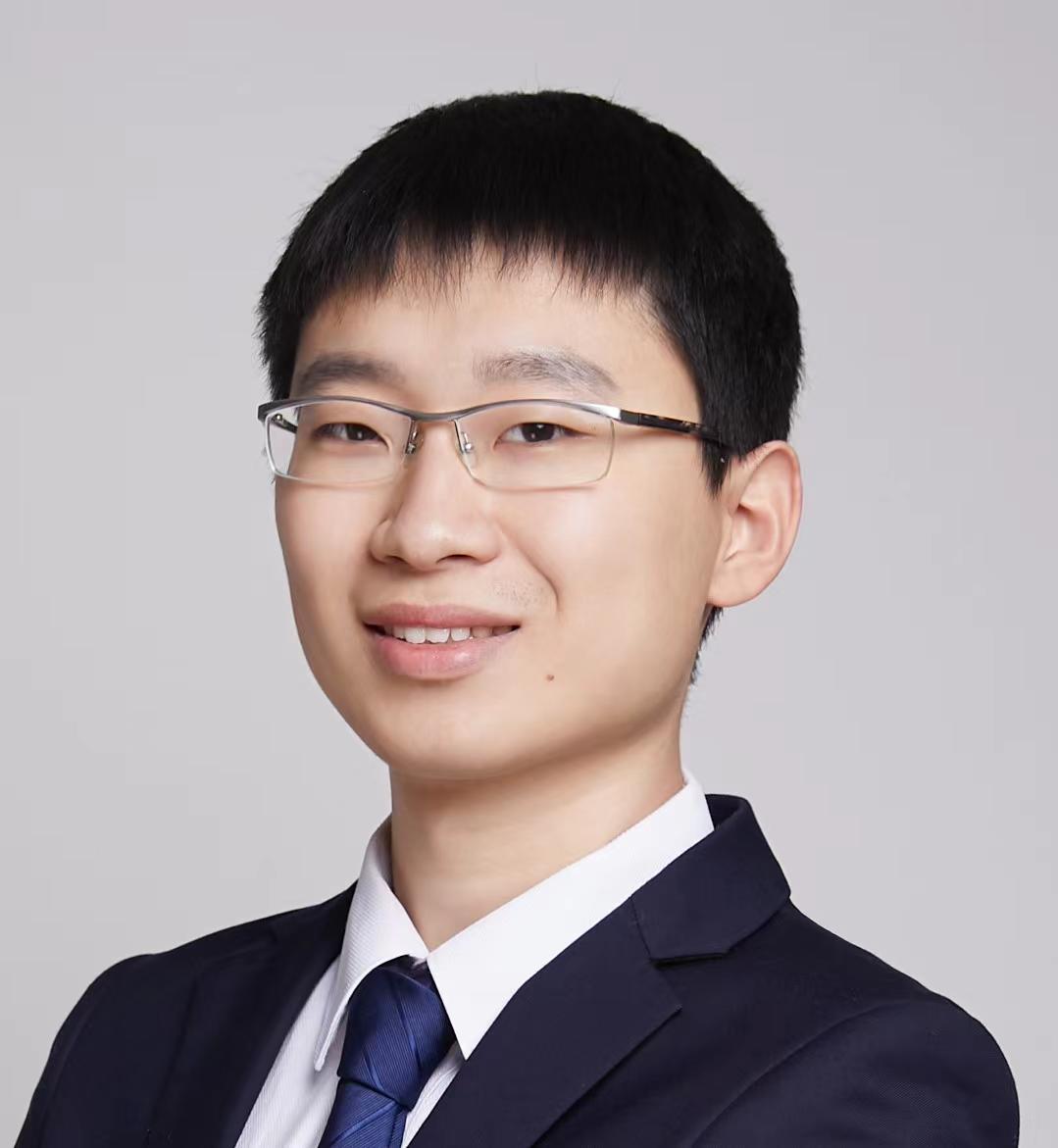}}]{Yuntian Chen} is an assistant professor at Eastern Institute of Technology, Ningbo. He received the B.S. degree from Tsinghua University, Beijing, China, in 2015, the dual B.S. degree from Peking University, Beijing, China, in 2015, and the Ph.D. degree with merit from Peking University, Beijing, China, in 2020. His research field includes scientific machine learning and intelligent energy systems. He
is interested in the integration of domain knowledge and data-driven models.
\end{IEEEbiography}

\begin{IEEEbiography}[{\includegraphics[width=1in,height=1.25in,clip,keepaspectratio]{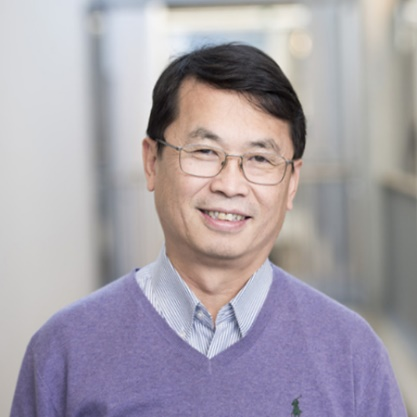}}]{Jerry Yan} is a member of the European Academy of Sciences and Arts and the Fellow of Hong Kong Academy of Engineering Sciences, currently serving as a Chair-Professor at the Hong Kong Polytechnic University. With a PhD from the Royal Institute of Technology (KTH), he has held chair professor positions at Luleå University of Technology, Mälardalen University, and KTH, Sweden. Prof. Yan's research focuses on renewable energy, advanced energy systems, climate change mitigation, and environmental policies. He has a publication record of over 500 papers in renowned journals such as Science, Nature Energy, Nature Climate Change etc and holds more than 10 patents. Having supervised nearly 200 post-doctoral researchers and 50 doctoral candidates, Prof. Yan has secured substantial external grants exceeding 20 million Euros. Prof. Yan's contributions have been acknowledged through prestigious awards, including the Global Human Settlements Award of Green Technology, the EU Energy Islands' Award, Research2Business Top100 and the IAGE Lifetime Achievement Award.
\end{IEEEbiography}


\end{document}


\title{Appendix}

\author{Dayin~Chen,
        Xiaodan~Shi{\IEEEauthorrefmark{1}},
        Haoran~Zhang,~\IEEEmembership{Senior Member,~IEEE,}       Xuan~Song{\IEEEauthorrefmark{1}},~\IEEEmembership{Member,~IEEE,}
        Dongxiao~Zhang, 
        Yuntian~Chen,~\IEEEmembership{Member,~IEEE,}
        and Jinyue~Yan{\IEEEauthorrefmark{1}}

\thanks{* Xiaodan Shi, Xuan Song and Jinyue Yan are the corresponding authors.}

\thanks{Dayin Chen is with Department of Building Environment and Energy Engineering, The Hong Kong Polytechnic University, Hong Kong SAR, China; International Centre of Urban Energy Nexus, The Hong Kong Polytechnic University,  Hong Kong SAR, China; Research Institute for Smart Energy, The Hong Kong Polytechnic University, Hong Kong SAR, China; Eastern Institute for Advanced Study, Eastern Institute of Technology, Ningbo, China. E-mail: 23038748r@connect.polyu.hk}
\thanks{Xiaodan Shi is with School of Business, Society and Technology, Mälardalens University, 72123 Västerås, Sweden; Center for Spatial Information Science, the University of Tokyo, 5-1-5 Kashiwanoha, Kashiwa-shi, Chiba 277-8568, Japan. E-mail: xiaodan.shi@mdu.se}
\thanks{Haoran Zhang is with School of Urban Planning and Design, Peking University, Shenzhen, China. E-mail: h.zhang@pku.edu.cn}
\thanks{Xuan Song is with School of Artificial Intelligence, Jilin University, Changchun, China; Research Institute of Trustworthy Autonomous Systems, Southern University of Science and Technology (SUSTech), Shenzhen, China. E-mail: songx@sustech.edu.cn}
\thanks{Dongxiao Zhang and Yuntian Chen are with Ningbo Institute of Digital Twin, Eastern Institute of Technology, Ningbo, China. E-mail: dzhang@eitech.edu.cn; ychen@eitech.edu.cn}
\thanks{Jinyue Yan is with Department of Building Environment and Energy Engineering, The Hong Kong Polytechnic University, Hong Kong SAR, China; International Centre of Urban Energy Nexus, The Hong Kong Polytechnic University,  Hong Kong SAR, China; Research Institute for Smart Energy, The Hong Kong Polytechnic University, Hong Kong SAR, China. E-mail: j-jerry.yan@polyu.edu.hk}

}




\maketitle

{\appendices

\section*{Feature Selection}


Previous studies frequently incorporate CPU utilization as one of the most essential input features \mbox{\cite{0,1}}. However, CPU utility is inaccessible on most phones now, rendering temperature measurement models based on CPU utility data of limited practical value. In the study \mbox{\cite{1}}, researchers also considered network connectivity as an input. However, the majority of mobile phones are consistently connected to Wi-Fi. Consequently, we do not employ CPU utility or network state as input features.

Furthermore, as indicated by the study \mbox{\cite{0}}, accurately measuring the heat production and dissipation rate of a mobile phone over a specific duration is crucial for estimating the ambient room temperature using the phone's battery temperature. Drawing upon this theory, we construct a novel feature set that effectively quantifies the rate of heat change in the mobile phone. For ease of reference, we provide a comprehensive list of all the features in Table \mbox{\ref{FEA_extra}}. In contrast to the features investigated in prior studies, these features possess a lower security level and are relatively easier to acquire.

\begin{table}[]
\center
\setlength{\tabcolsep}{4.5mm}
\caption{The collected features from smartphones.}
\label{FEA_extra}
\begin{tabular}{ll}
\hline
Symbol & Collect Features\\ \hline
$F_1$ & Screen state\\
$F_2$& Battery Voltage \\
$F_3$ & Battery Temperature\\
$F_4$&Continuous screen activation time\\
$F_5$&Continuous screen off time\\
$F_6$&Before active, the continuous screen off time\\
$F_7$&Before off, the continuous screen active time\\
$F_8$&Final Battery temperature when screen was activated \\
$F_9$& Final Battery temperature when screen was turn off \\ \hline
\end{tabular}
\center
\end{table}

The first feature, denoted as $F_1$, is a binary value indicating the activation status of the phone screen. Features $F_2$ and $F_3$ represent the voltage and temperature of the phone battery, respectively, which are directly obtained. To provide a clearer understanding of the remaining features, $F_4$ to $F_9$, we utilize Fig. \mbox{\ref{fea_illustration}} to illustrate them. We present two examples of phones in different states: one with an active screen and the other with the screen off. In Fig. \mbox{\ref{fea_illustration}}, the battery temperature is denoted as $T$, and $t$ represents the duration of time.

\begin{figure*}
	\centering
		\includegraphics[scale=0.5]{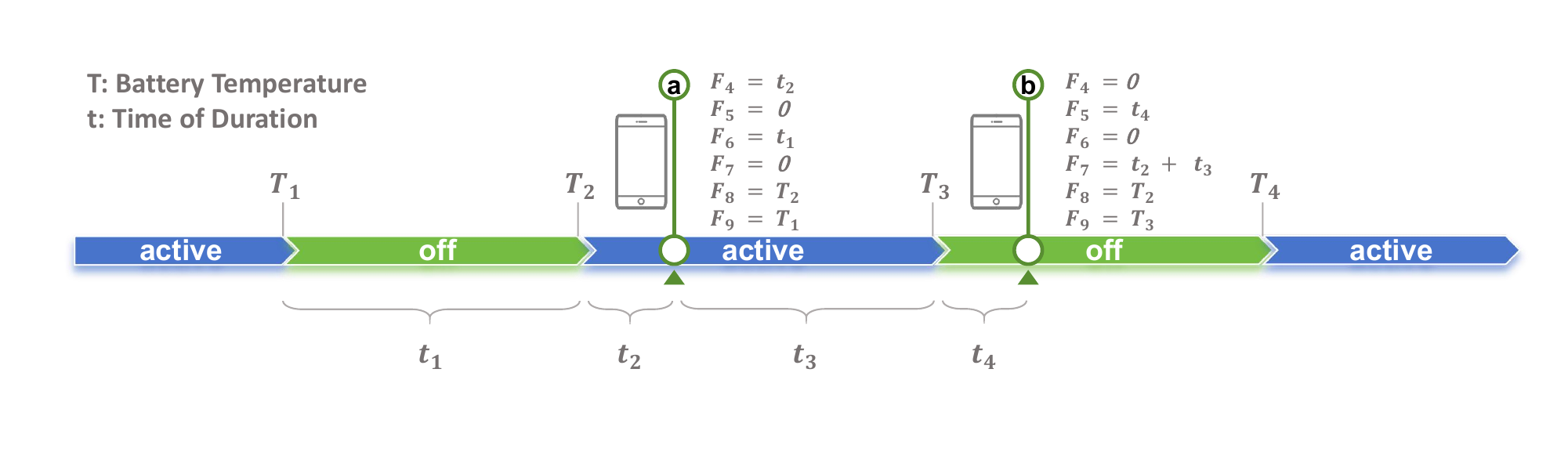}
	\caption{The illustration of what the features $F_4$ to $F_9$ represent for.
}
	\label{fea_illustration}
\end{figure*}

\section*{Phone Context Setting}

Taking into account the impact of indoor environmental characteristics and various phone contexts (such as being in the pocket, on the desk, or in the hand), we design that our system activates the ambient temperature estimation model only when the phone is detected to be exposed to the air. This strategy is also used in the work \mbox{\cite{1}}. For a more comprehensive illustration, we refer readers to our previous work \mbox{\cite{chen2023}}.

\section*{Design of Loss Function}

In Fig. \ref{gaussian}, we illustrate our design by presenting three distinct predicted distributions. As inherent to the Gaussian distribution, increased variance results in a wider range of potential values and reduces the likelihood of observing values close to the expectation. That means the true temperature value possesses a greater probability of deviating from the expectation. Conversely, a smaller variance leads to a more concentrated distribution, thereby indicating that the ground truth is less likely to deviate significantly from the expected value. This is precisely why variance is used to denote the uncertainty of the prediction.

\begin{figure}
	\centering
		\includegraphics[scale=0.4]{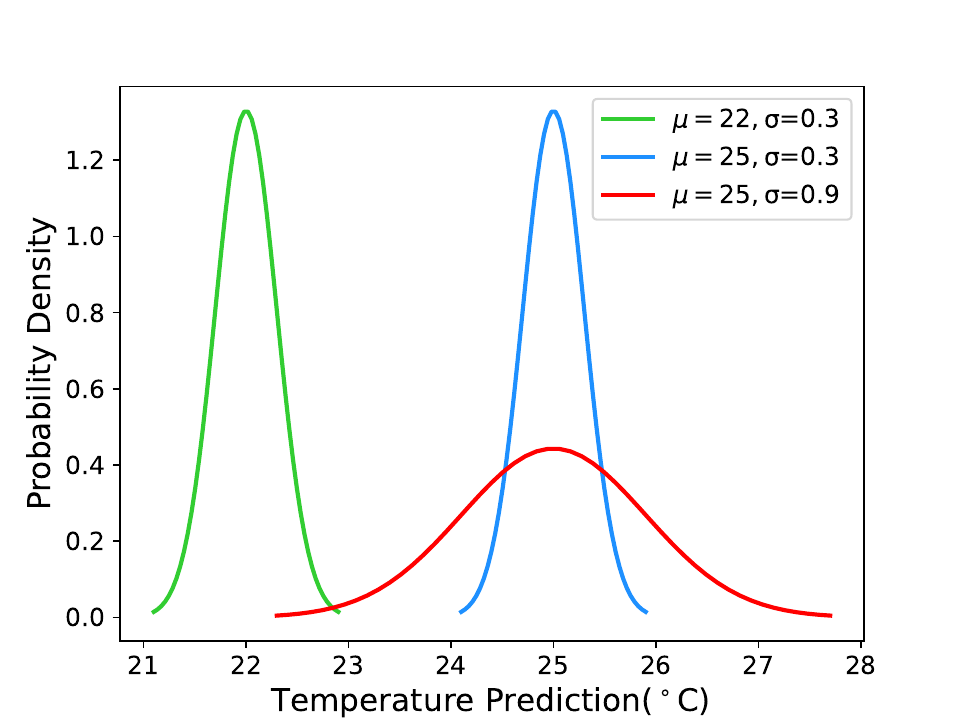}
	\caption{Demonstration of different predicted distributions. The $\mu$ in every distribution indicates the estimation ambient temperature values while the $\sigma$ indicates the uncertainty. We can see with the increasing of uncertainty, the temperature will have a wider range of possible values and it also will have a smaller possibility of locating at $\mu$.
}
	\label{gaussian}
\end{figure}

\section*{Effectiveness of Uncertainty Value}

To provide a more intuitive representation of the effectiveness of uncertainty values, we plot the relationship between uncertainty and estimation bias using data samples from the validation data set of each contributor. By examining the results depicted in Fig. \ref{uncertainty}, it becomes evident that as the uncertainty value increases, the distribution of prediction bias becomes more divergent. This indicates that uncertainty is helpful for evaluating answers in the crowdsourcing process

\begin{figure*}
    \centering
  \subfloat[$C_1$]{\includegraphics[width=2.3in]{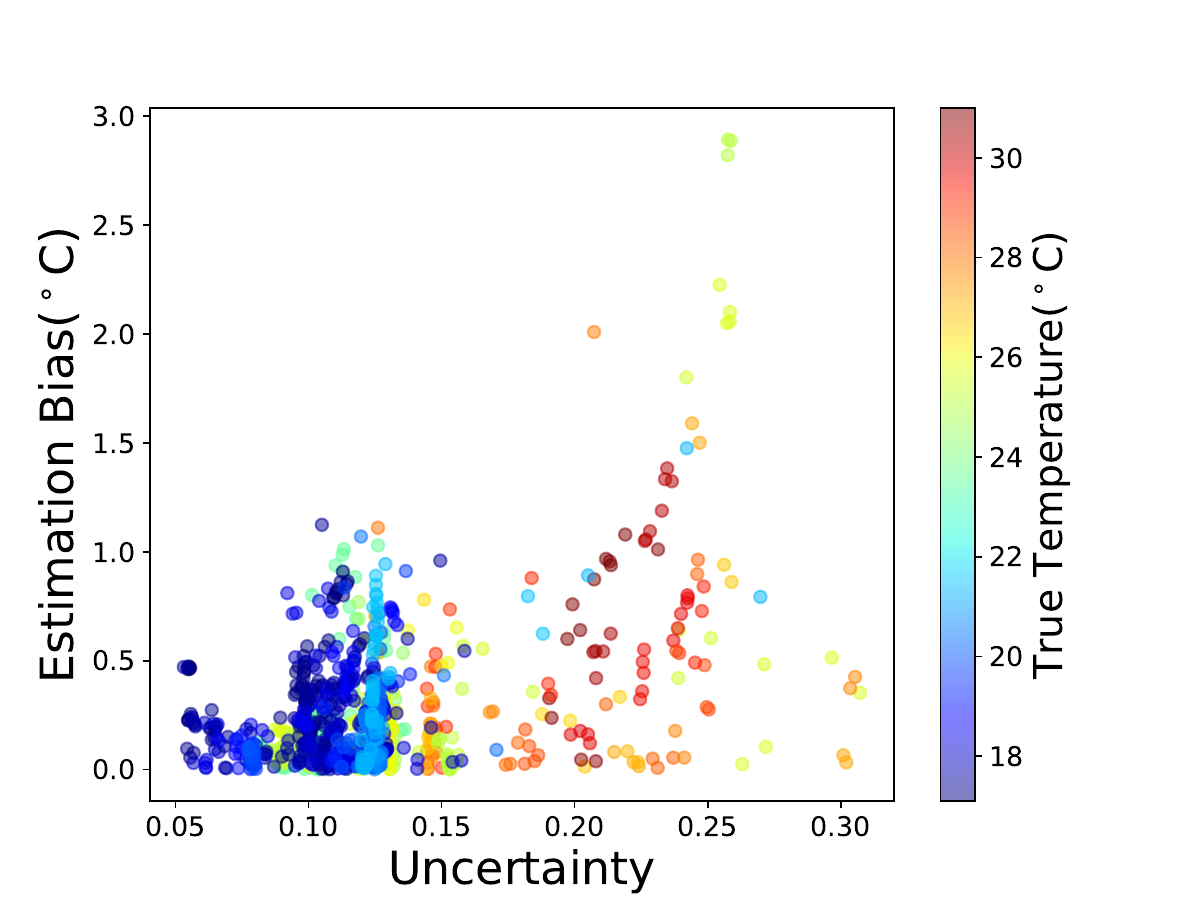}}
  \hfil
  \subfloat[$C_2$]{\includegraphics[width=2.3in]{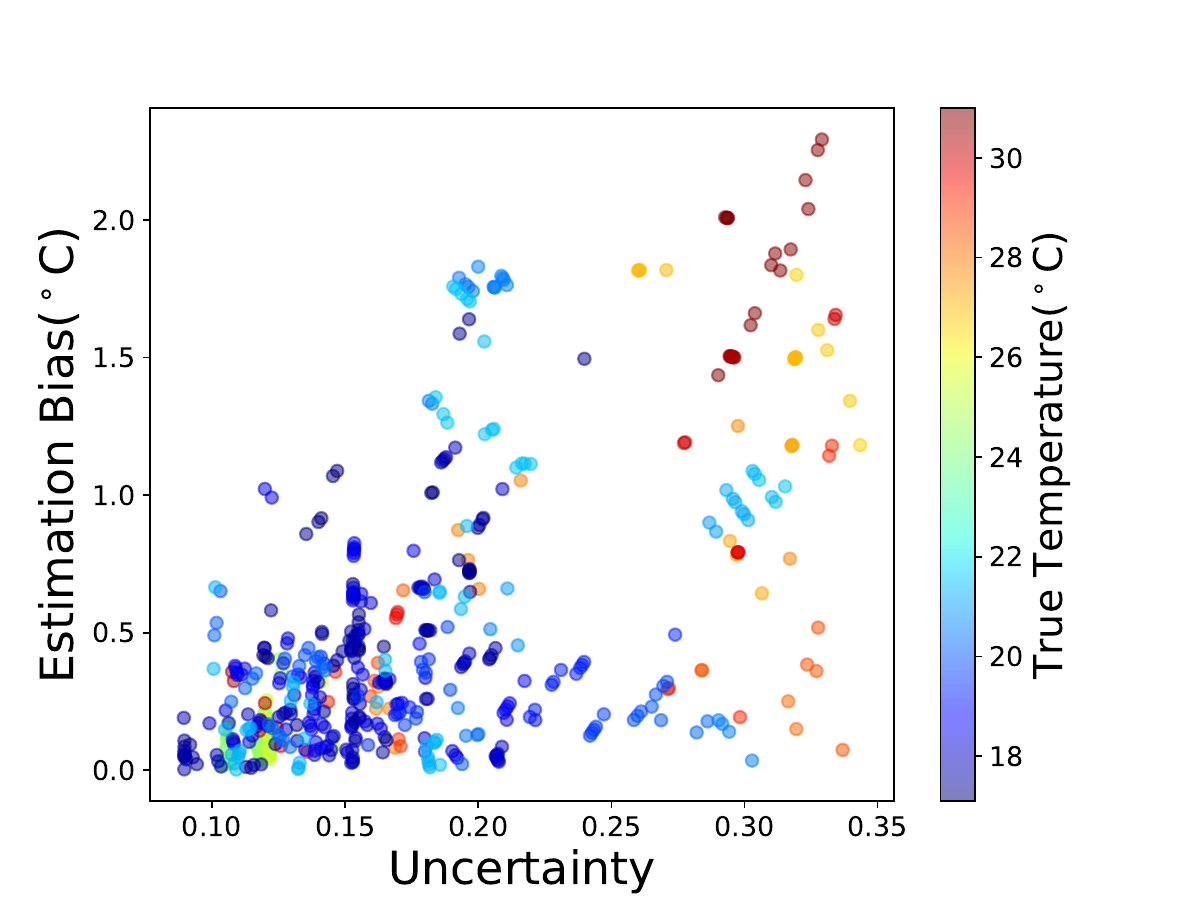}}
  \hfil
  \subfloat[$C_3$]{\includegraphics[width=2.3in]{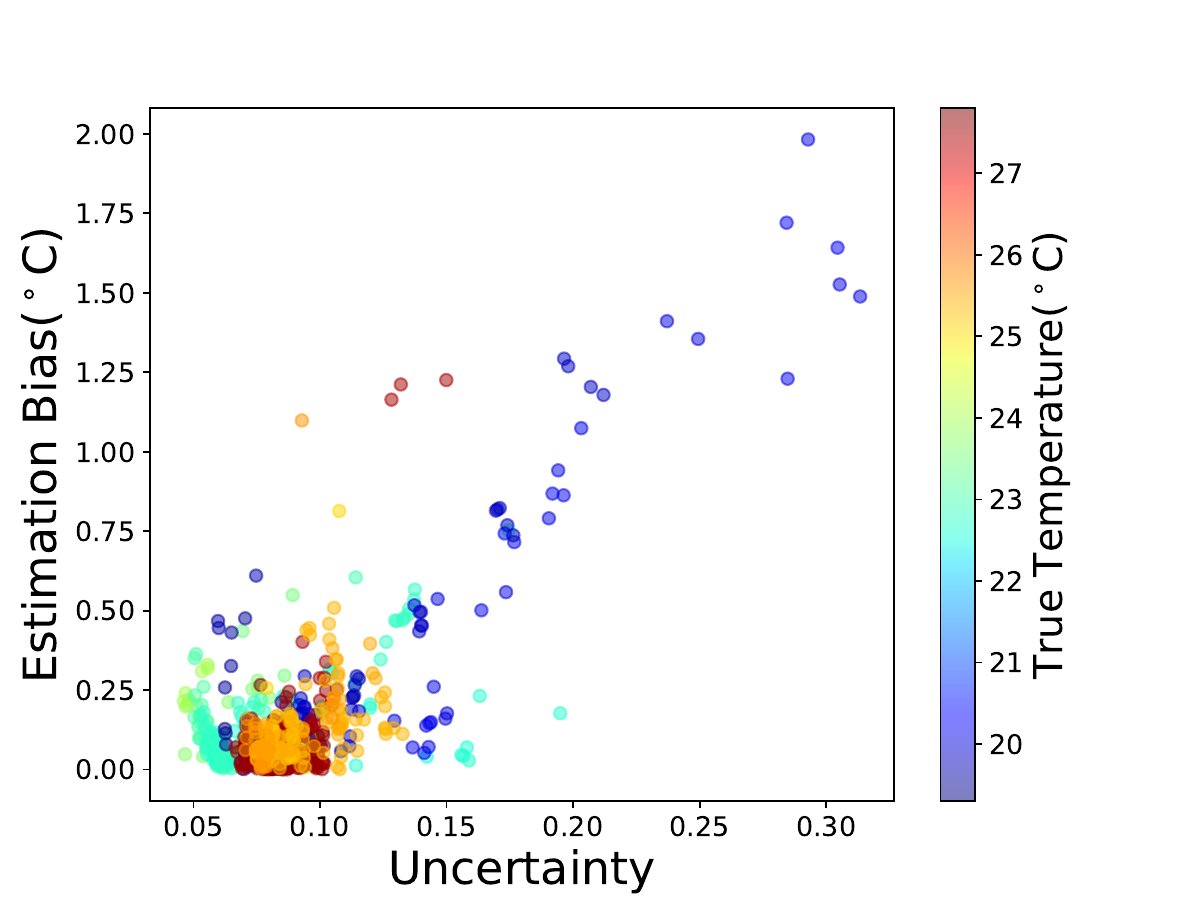}}\\

  \subfloat[$C_4$]{\includegraphics[width=2.3in]{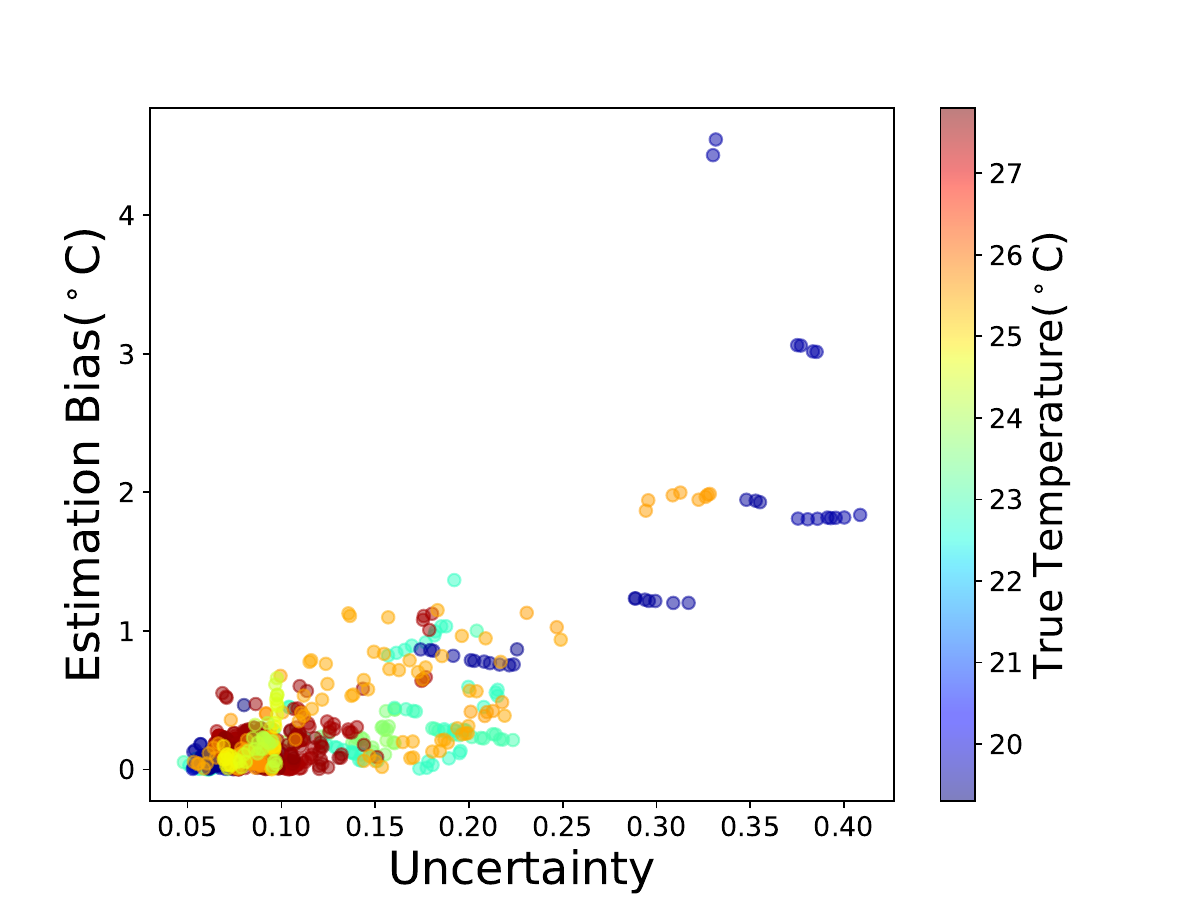}}
  \hfil
  \subfloat[$C_5$]{\includegraphics[width=2.3in]{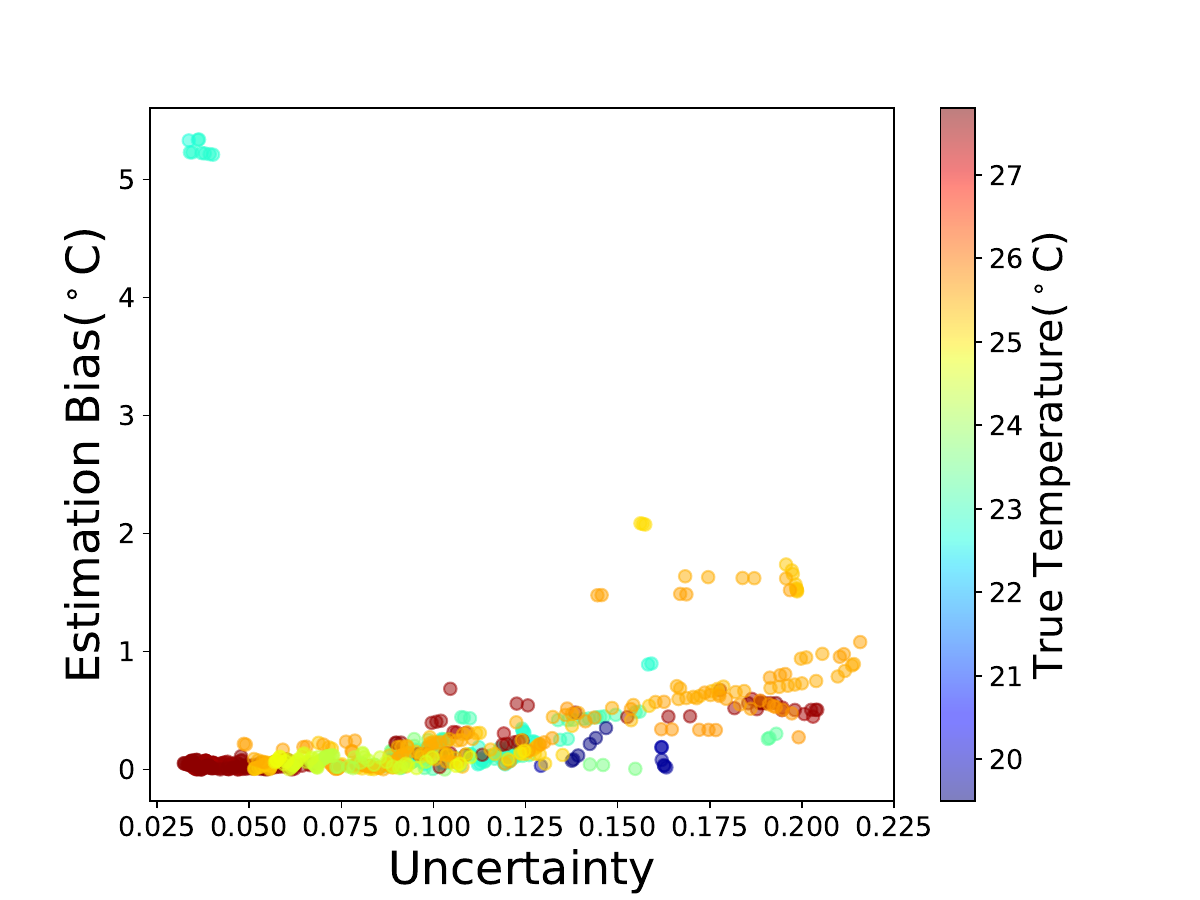}}
  \hfil
  \subfloat[$C_6$]{\includegraphics[width=2.3in]{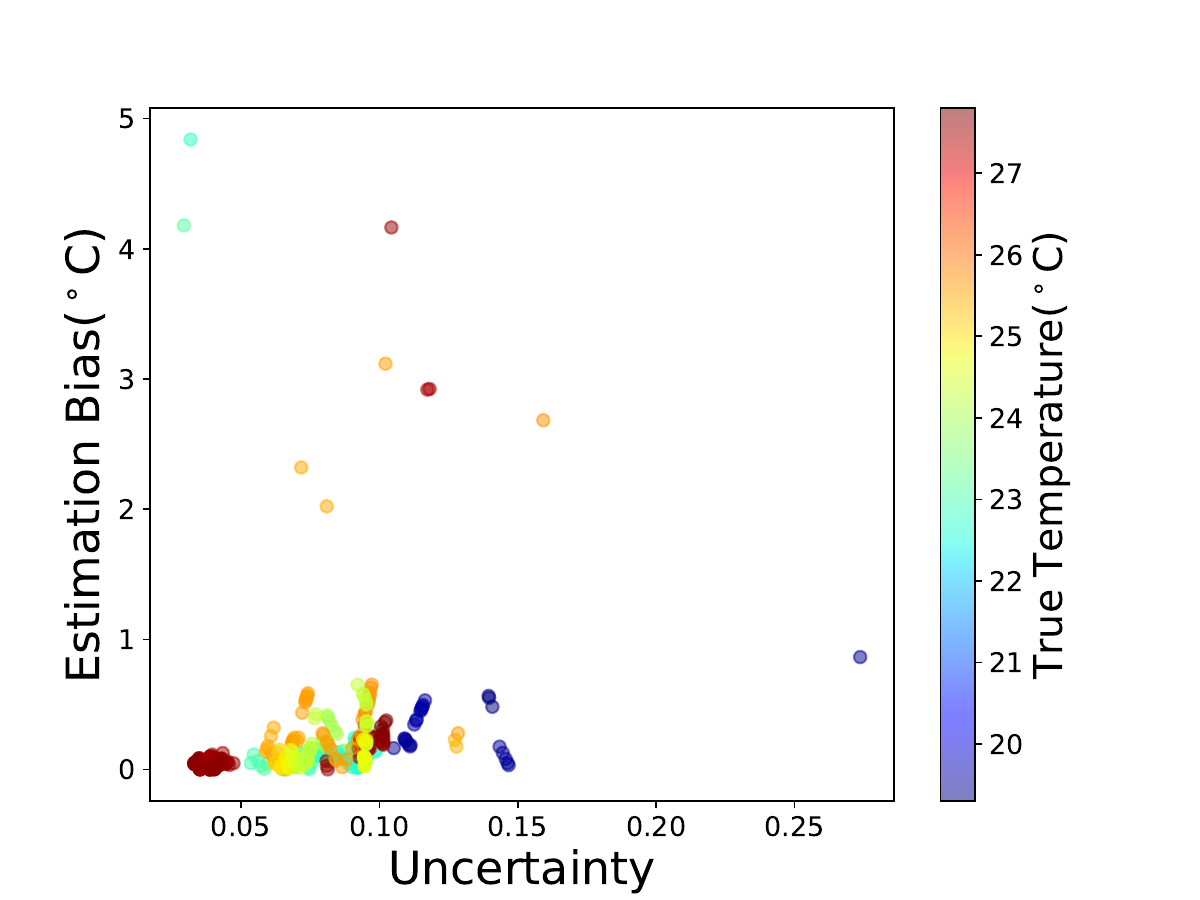}}
  \caption{The relationship diagrams depicting the uncertainty and estimation bias which is constructed using the validation data set obtained from all contributors. In each diagram, the horizontal axis represents the level of uncertainty, while the vertical axis portrays the estimation bias. The colors utilized in the diagrams indicate the true temperature values associated with each piece of data.}
  \label{uncertainty}
\end{figure*}

\section*{Crowdsourcing Truth Inference Baselines Description}

We select the crowdsourcing baseline methods from two research areas to compare with our CBTS model:
\begin{itemize}
  \item \textbf{Truth Inference}: We retrieve various truth inference methods used in crowdsourcing between 1979 and 2023. However, most of these methods are designed for decision-making tasks \mbox{\cite{GLAD, min&max, BCC, CBCC, LFC, multi, KOS, VIBP}} or single-choice/multi-choice tasks \mbox{\cite{uncertainty3, uncertainty4, uncertainty5,GLAD, min&max, BCC, CBCC, LFC}}, and are not directly applicable to numeric tasks. In our study, we select four representative truth inference methods (Mean \mbox{\cite{cs_all}}, D\&S \mbox{\cite{DS}}, PM \mbox{\cite{PM}}, ZC \mbox{\cite{ZC}}) that are suitable for numeric tasks to compare with our CBTS truth inference model. Additionally, we include the Majority Voting (MV) method \mbox{\cite{cs_all}} as another baseline and apply a minor adjustment to tailor it for our purposes. Although this method is not originally designed for numeric tasks, it has demonstrated strong performance and thus warrants consideration in our evaluation. Specifically, we cluster all the answers and select the average of the largest cluster as the result. We conduct evaluation experiments with two different clustering numbers: 2 and 3. In cases where the number of given answers is smaller than the clustering number, we treat them as a single cluster and directly compute the average as the result.

  \item \textbf{Phone-based Ambient Temperature Estimation}: We reviewed existing studies on phone-based ambient temperature estimation that utilize crowdsourcing technology \mbox{\cite{1,2,3,4}}. In the works \mbox{\cite{2,3,4}}, researchers crowdsource the temperature values by calculating the average, indicating the utilization of the Mean \mbox{\cite{cs_all}} method for truth inference. While in the work \mbox{\cite{1}}, the researchers employed a weighted average approach, considering the confidence level of each phone. In order to reproduce this method, we assigned a confidence value to each phone based on its performance on the training data and adopted the weighted average method to crowdsource the answers. This method is referred to as Weighted Average (WA). In this study, we denote the MAE on the training data set of phone $C_n$ as $E_n$. We calculate the weight $W_n$ for each phone using the following formula:
  \begin{equation}
    W_n = \frac{\frac{1}{E_n}}{\sum_{i=1}^6\frac{1}{E_i}}
  \label{XX2}      
  \end{equation}
  The final weights obtained for the six phones are as follows: $W_1 = 0.144$, $W_2 = 0.08$, $W_3 = 0.262$, $W_4 = 0.141$, $W_5 = 0.136$, $W_6 = 0.237$.

\end{itemize}

\bibliographystyle{ieeetr}
\bibliography{sample-base}